\documentclass[letterpaper]{article} % DO NOT CHANGE THIS
\usepackage[submission]{aaai2026}  % DO NOT CHANGE THIS
\usepackage{times}  % DO NOT CHANGE THIS
\usepackage{helvet}  % DO NOT CHANGE THIS
\usepackage{courier}  % DO NOT CHANGE THIS
\usepackage[hyphens]{url}  % DO NOT CHANGE THIS
\usepackage{graphicx} % DO NOT CHANGE THIS
\usepackage{natbib}  % DO NOT CHANGE THIS AND DO NOT ADD ANY OPTIONS TO IT
\usepackage{caption} % DO NOT CHANGE THIS AND DO NOT ADD ANY OPTIONS TO IT
\usepackage{enumitem}
\usepackage{algorithm}
\usepackage{algorithmic}
\usepackage{newfloat}
\usepackage{listings}

\usepackage{xargs}   
\usepackage{xspace}

\usepackage{multirow}
\usepackage{verbatim}
\usepackage{xcolor}

\usepackage{array}
\usepackage{amssymb}
\usepackage{bm}
\usepackage{amsmath}

\usepackage{multirow}
\usepackage{booktabs}

\usepackage{nameref}

\makeatletter
\def\showauthors@on{T}
\makeatother

\DeclareCaptionStyle{ruled}{labelfont=normalfont,labelsep=colon,strut=off} % DO NOT CHANGE THIS
\floatstyle{ruled}
\newfloat{listing}{tb}{lst}{}
\floatname{listing}{Listing}
\title{\texttt{ForeSWE}: Forecasting Snow-Water Equivalent with an Uncertainty-Aware Attention Model}
\author{
    Krishu K Thapa,
    Supriya Savalkar,
    Bhupinderjeet Singh,
    Trong Nghia Hoang,
    Kirti Rajagopalan,
    Ananth Kalyanaraman
}
\affiliations{
        Washington State University, Pullman, WA.\\
    \{krishu.thapa, supriya.savalkar, bhupinderjeet.singh, trongnghia.hoang, kirtir, ananth\}@wsu.edu
}

\definecolor{darkgrn}{rgb}{0, 0.75, 0}
\newcommand{\modified}[1]{\textcolor{darkgrn}{#1}}

\usepackage[textsize=scriptsize,textwidth=25mm]{todonotes}

\newcommandx{\nh}[2][1=]{\todo[color=green!50,#1]{\sf \textbf{Nghia:} #2}\xspace}

\newcommandx{\ak}[2][1=]{\todo[color=green!50,#1]{\sf \textbf{AK:} #2}\xspace}

\newcommandx{\kt}[2][1=]{\todo[color=blue!50,#1]{\sf \textbf{KT:} #2}\xspace}

\newcommandx{\kr}[2][1=]{\todo[color=green!50,#1]{\sf \textbf{Kirti:} #2}\xspace}

\newcommandx{\bs}[2][1=]{\todo[color=green!50,#1]{\sf \textbf{Bhupi:} #2}\xspace}

\newcommandx{\supriya}[2][1=]{\todo[color=green!50,#1]{\sf \textbf{Supriya:} #2}\xspace}

\newcommand{\snotel}{SNOTEL}

\newcommand{\tpatt}{\texttt{Tp-Att}}
\newcommand{\spatt}{\texttt{Sp-Att}}
\newcommand{\lstm}{\texttt{LSTM}}
\newcommand{\rgp}{\texttt{Raw-GP}}
\newcommand{\tar}{\texttt Transformer}
\newcommand{\nva}{\texttt{NVA-Base}}

\newcommand{\fs}{\texttt{ForeSWE}}
\newcommand{\tft}{\texttt{TFT}}

\DeclareMathOperator*{\argmax}{\text{argmax}}

\begin{document}

\maketitle

\begin{abstract}
    Various complex water management decisions are made in snow-dominant watersheds with the knowledge of Snow-Water Equivalent (SWE)---a key measure widely used to estimate the water content of a snowpack. However, forecasting SWE is challenging because SWE is influenced by various factors including topography and an array of environmental conditions, and has therefore been observed to be spatio-temporally variable. Classical approaches to SWE forecasting have not adequately utilized these spatial/temporal correlations, nor do they provide uncertainty estimates---which can be of significant value to the decision maker. In this paper, we present \fs{}, a new probabilistic spatio-temporal forecasting model that integrates deep learning and classical probabilistic techniques. The resulting model features a combination of an attention mechanism to integrate spatiotemporal features and interactions, alongside a Gaussian process module that provides principled quantification of prediction uncertainty.
We evaluate the model on data from 512 Snow Telemetry (SNOTEL) stations in the Western US. The results show significant improvements in both forecasting accuracy and prediction interval compared to existing  approaches. The results also serve to highlight the efficacy in uncertainty estimates between different approaches. Collectively, these findings have provided a platform for deployment and feedback by the water management community.

\end{abstract}

\section{Introduction}
\label{sec:intro}

Streamflow is vital for societal needs, including food production, irrigation, flood control, hydropower, and supporting endangered fish species. In snow-dominant watersheds of the Western U.S., 50–80\% of annual streamflow originates from melting winter snowpack \citep{hunter2006oceanic, li2017much}. Therefore, the state of the snowpack and the water it contains—known as \emph{Snow Water Equivalent (SWE)}—is critical for determining the magnitude and timing of streamflow \citep{mankin2015potential, harpold2017rain}.

Information on the current state of SWE is widely used by local and federal water agencies in the United States, including reservoir operators and irrigation districts \citep{usbr_snowwater_supply_forecasting}. The National Resources Conservation Service (NRCS) maintains a nationwide dashboard for monitoring SWE and related variables \citep{usda_nrcs_snow_water_interactive_map}. However, SWE forecast information, while critical, is not yet available in an operational context.

Accurate SWE forecasts are critical for both short- and long-term water management. Forecasts at multi-day, and multi-week timescales each serve important roles: multi-day forecasts help anticipate rapid snowmelt and potential flooding, enabling real-time interventions such as reservoir drawdowns, while multi-week forecasts of SWE and peak SWE inform subseasonal planning and allocation decisions across agricultural, ecological, and hydropower sectors \citep{pagano2004evaluation, huang2017evaluation, stillinger2021reservoir}. SWE also serves as a key input to streamflow forecasts \citep{mote2005declining} and improves sub-seasonal climate outlooks by capturing important land–atmosphere feedbacks \citep{subseasonalForecast}.

SWE forecasts can contribute to the growing body of climate-informed tools that offer direct potential for social good. Improvements in SWE forecasting can support more equitable and sustainable water use, reduce the societal impacts of extreme hydrologic events, and strengthen the resilience of communities and ecosystems that depend on predictable water availability. %These outcomes are especially critical in regions already facing the compounding pressures of climate change, competing water demands, and environmental degradation.

\paragraph{\bf Challenges.}~Real-time forecasting of SWE as the season progresses, is challenging due to variations in SWE patterns across space and time. These patterns are  influenced by complex interactions between local weather and spatial attributes, as well as by the phase of SWE. %This results in different snowmelt effects depending on the phase of SWE and  environmental conditions.

\begin{itemize}[leftmargin=5pt]\itemsep=-0.05ex
    \item {\bf Temporal Variation.}~For any location, the state of its snowpack at any point influences its SWE trend in the future. For example, a dry and cold snowpack has more liquid retention capacity and can act as buffer to reduce the risk of melt and flooding during storm events \citep{GarvelmannPohlWeiler2015_110102340}. 
    %Similarly, snow-depth also matters \citep{Wrzer2016InfluenceOI}. 
    Meteorological variables (e.g., temperature, precipitation) vary temporally and determine the timing of snow accumulation and melt.
    %both the typical seasonal pattern of SWE (e.g; when snow accumulates or starts to melt), and variations around it.
    
    %Capturing such temporal patterns and its impact is therefore essential to improve forecasting accuracy.
    
    \item {\bf Spatial Variation.}~ 
    The variation of SWE across locations results from different interrelated factors like orographic effects, elevation, terrain, wind, vegetation and radiation \citep{liston_glen}. For instance,  locations in higher elevation can have colder temperature and more snow accumulation; and southern-facing slopes can have faster snowmelt due to solar exposure. 
    %c) The energy from the longwave radiation emitted by vegetation can affect snow retention and  melt rates. 
    Additionally, locations at different ranges of proximity might exhibit spatial correlations (e.g.,  areas falling under same atmospheric river path). %These examples are not prescriptive as interactions across factors matter, but capturing such spatial variations and correlations will help generalize the forecast reliably across locations. 
    
    \item {\bf Spatio-Temporal Interaction.}~Spatiotemporal attributes of a location interact and influence SWE behavior \citep{GarvelmannPohlWeiler2015_110102340}. For example, the relationship between surface air temperatures and precipitation phase (rain or snow) depends on relative humidity \citep{jennings}. 
    The precipitation phase can in turn impact snowmelt, creating a potential for  floods \citep{musselman}.  
    %How the precipitation phase impacts snowmelt can depend on the state of snowpack. 
   % Learning these interactions is essential for an accurate forecast.
    %Learning such varying spatio-temporal interactions is essential to improve SWE forecasting.

    \item {\bf Environmental Uncertainty.}~SWE trends are affected by the inherent stochastic nature of the process and unpredictable environmental anomalies. For example, an ongoing accumulation phase might unexpectedly shift to rapid melt due to untimely warming. Accounting for such uncertainty in the SWE forecasting is essential to optimize for planning decisions regarding resource deployment.
    
\end{itemize}

\captionsetup{font={normalsize}}
\begin{figure}[tb]
\centering
\includegraphics[scale=0.34]{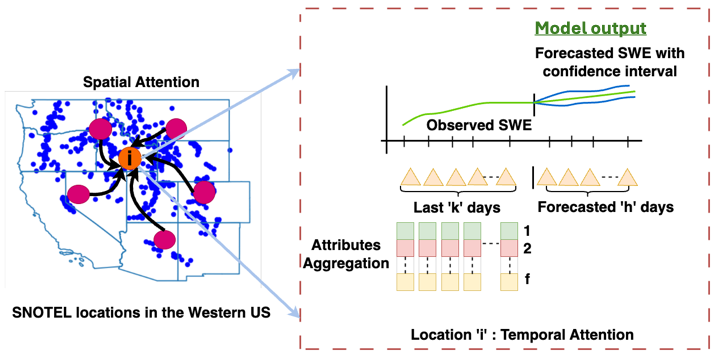}
\captionof{figure}{\normalsize Spatiotemporal SWE forecasting with confidence interval for  location $i$ over a horizon $h$, using $k$ days of historical observations with $f$ attributes. 
Each blue dot in the map is a SNOTEL location. 
%Shown left is the main idea behind spatial attention relative to a target location ($i$) with the blue dots denoting all locations with SNOTEL stations. Shown right is the deep temporal view within a location along with its attribute space in our forecasting model. 
}
\label{fig:intro_fig}
\end{figure}

\vspace*{-0.05in}
\paragraph{\bf Contributions.} 
In this paper, we present \fs{}, a new uncertainty-aware attention model for SWE forecasting, that combines modern deep learning and classical probabilistic methods (see Figure~\ref{fig:intro_fig}). 
% We target both daily and weekly scales for forecasting, because both are used for different types of decision making: daily for emergency response situations such as flood risks, and weekly for irrigation decisions in agriculture, recreational planning (e.g., skiing), and energy (e.g., hydropower generation).  
We focus on SWE forecasting at both daily and weekly timestep, as each serves distinct decision-making needs: daily forecasts are critical for emergency response scenarios such as flood risk management, while weekly forecasts inform  agricultural planning, recreational activities (e.g., skiing season operations) and energy management (e.g., hydropower generation).

The paper makes the following  contributions:\vspace{0.5mm}

{\noindent \bf 1.} An attention-based deep learning (DL) model~\citep{vaswani2017} parameterized with a new spatio-temporal attention module which is specifically designed to capture the correlation in space and time, as well as interaction between attributes in SWE forecasting context.

{\noindent \bf 2.} 
% A probabilistic augmentation which replaces the prediction head of the above DL model with a Gaussian process (GP)~\citep{Rasmussen06}---providing a principled framework for learning spatio-temporal correlation and computing uncertainty quantification for SWE forecasting.
A probabilistic augmentation using a Gaussian process (GP)~\citep{Rasmussen06} as the prediction head, enabling spatio-temporal correlation learning and uncertainty quantification of SWE forecasts.

{\noindent \bf 3.}
A standalone sparse raw GP implementation that represents a GP-only baseline for the SWE forecasting problem.

{\noindent \bf 4.} 
% An extensive empirical study to evaluate the improved forecasting efficacy and prediction interval of the proposed model over previous approaches, with results showing that our daily forecasting model outperforms all state-of-the-art models for $90\%$ of the locations with a minor deviation from optimal performance for the rest of the locations.
A thorough experimental evaluation that evaluates our proposed approaches against various spatial and/or temporal ML approaches. 
Results show that for daily forecasting \fs{} and \rgp{} outperform all other approaches, while for weekly forecasting, \fs{} is the most suited model as it delivers the best performance both by accuracy and by the quality of its uncertainty estimates.

%An empirical evaluation demonstrating that the proposed model outperforms state-of-the-art methods for $90\%$ of locations, with minimal deviations for the remaining locations.

\section{Related Work}
\label{sec:related_work}

The current literature on SWE forecasting features both mechanistic and data-driven approaches. Mechanistic models utilize prior descriptive knowledge in physical and hydrological equations. However, our knowledge of these processes is limited leading to simplified models and large biases  ~\citep{subseasonalForecast}. Alternatively, machine learning (ML) techniques can learn from historical data from a diverse collection of locations, and benefit where physical knowledge may be lacking; and be learnable in an incremental fashion as new data become available---e.g., past temporal observations of SWE,  and additional spatial features such as remotely-sensed reflectances~\citep{brodzik2016measures}. 
%For examples, most recent practical forecasting models have leveraged data collected from $512$ snow telemetry (SNOTEL) stations in Western US for a data-driven approach 
%\bs{citation missing?}.\KT{Is this talking about our previous paper? Cause its prediction and not forecasting and has 323 stations} 
%This includes (past) temporal observations of SWE and additional spatial features such as remotely-sensed reflectances~\citep{brodzik2016measures}, topographical details, and meteorological data.
Such data can be utilized to build SWE forecasting models.
%which, for example, provide a 10-day forecast for short-term planning and a 4-week forecast for long-term strategies. 
%ML models can be adapted more flexibly to new locations and weather conditions via updating its parameters with new data, which presents a more scalable approach. 
% However,  existing ML-based forecasting models ~\citep{sarhadi2014snow,CUI2023128835,thapa2024attention,FRANZ2010820} do not explicitly model spatio-temporal interactions or correlations or provide limited to no uncertainty estimates for their prediction,  which are essential for  downstream decision making.  %among spatial and temporal attributes as well as their correlation across different locations and time. Furthermore, prior approaches have also not provided principled uncertainty quantification of their SWE forecasting which is essential for optimizing downstream decision making.

Among the earlier data-driven works in SWE forecasting, \citep{sarhadi2014snow} uses statistical models like ARIMA \citep{arima} and SARIMA \citep{sarima} with exogenous variables to predict daily (i.e. 30 days) and monthly (i.e. 6 months) SWE. % in Ontario, Canada. 
%These models feed the daily or monthly historical SWE trend with external factors or variables that can impact the SWE in a given location and time, thereby forecasting the SWE value. 
However, all the locations under study are in low elevation ($<$2000ft) with a low annual max SWE (100 to 150mm), making the model too restrictive. Similarly, \citep{FRANZ2010820} combines twelve bio-physical models using a Bayesian Model Averaging (BMA) to forecast SWE with uncertainty quantification. Although the work shows competitive results,   it has only been implemented on six locations and limited to a single day forecasting. 
%have a limited forecasting window of one day.
%, unlike our dataset with locations at varying elevations (2000 to 11000 ft) and annual max SWE (50 to 2000mm), adding variability and challenges to model building}.  
% \citet{subseasonalForecast} focus on the sub-seasonal forecasting (h = 32 daily or 4 weekly points forecasts) of gridded SWE worldwide using an ensemble of existing global physical and hydrological models such as GEPS \citep{charron2010toward, houtekamer2009model}, GEFS \citep{toth1993ensemble, zhu2017impact, zhu2018toward, li2019evaluating}, and FIM \citep{bleck2010use, lee2009finite}.
% This approach relies on the equations and assumptions of all the underlying models---making it difficult to incorporate new factors and recalibrate. Additionally, over the years assumptions are subjected to change with the evolving climate and weather patterns.

Deep Learning approaches have been explored more recently. \cite{CUI2023128835} couples the Long Short-Term Memory (LSTM) \citep{hochreiter1997long} and zonal bias correction data assimilation approach to predict the gridded SWE value at 1 km pixel for the next day or month. 
%Here, LSTM captures temporal correlations, and the zonal bias correction performs inverse-weighted assimilation of SWE from the observation sites in the surroundings. 
The assimilation approach depends on the presence of nearby observation sites, which makes it challenging due to the geographical sparsity in the observation layer. %Moreover, it assumes spatial correlation is proportional to proximity, which is restrictive  (e.g., locations falling under the same atmospheric river path can be correlated, although far away).
%neglecting the fact that locations that might not necessarily be nearby can still have strong correlations. 
%For example, locations falling under the same atmospheric river path, although far away can have similar behavior. 
Recently, an attention-based model was implemented to predict SWE using spatial and temporal attention \citep{thapa2024attention} but it does not provide any uncertainty estimates. For a different application in climate modeling, a few approaches exist \citep{nguyen2023climax, nguyen2023scaling}.
 In comparison to these approaches, our model architecture is different---as it uses a form of spatio-temporal attention that  captures attribute interaction  %through an aggregation, making the query vector in its attention mechanism vary with location and time 
 (unlike a single learnable query vector in the \textit{ClimaX model} \citep{nguyen2023climax}), and is designed to provide  uncertainty estimates.

\section{Model Architecture}
\label{sec:model_arch}

\captionsetup{font={normalsize}}
\begin{figure*}[tb]
    \begin{minipage}{\textwidth}
        \centering
        \vspace{0pt}
        \begin{minipage}{0.45\textwidth} 
            \includegraphics[scale=0.32]{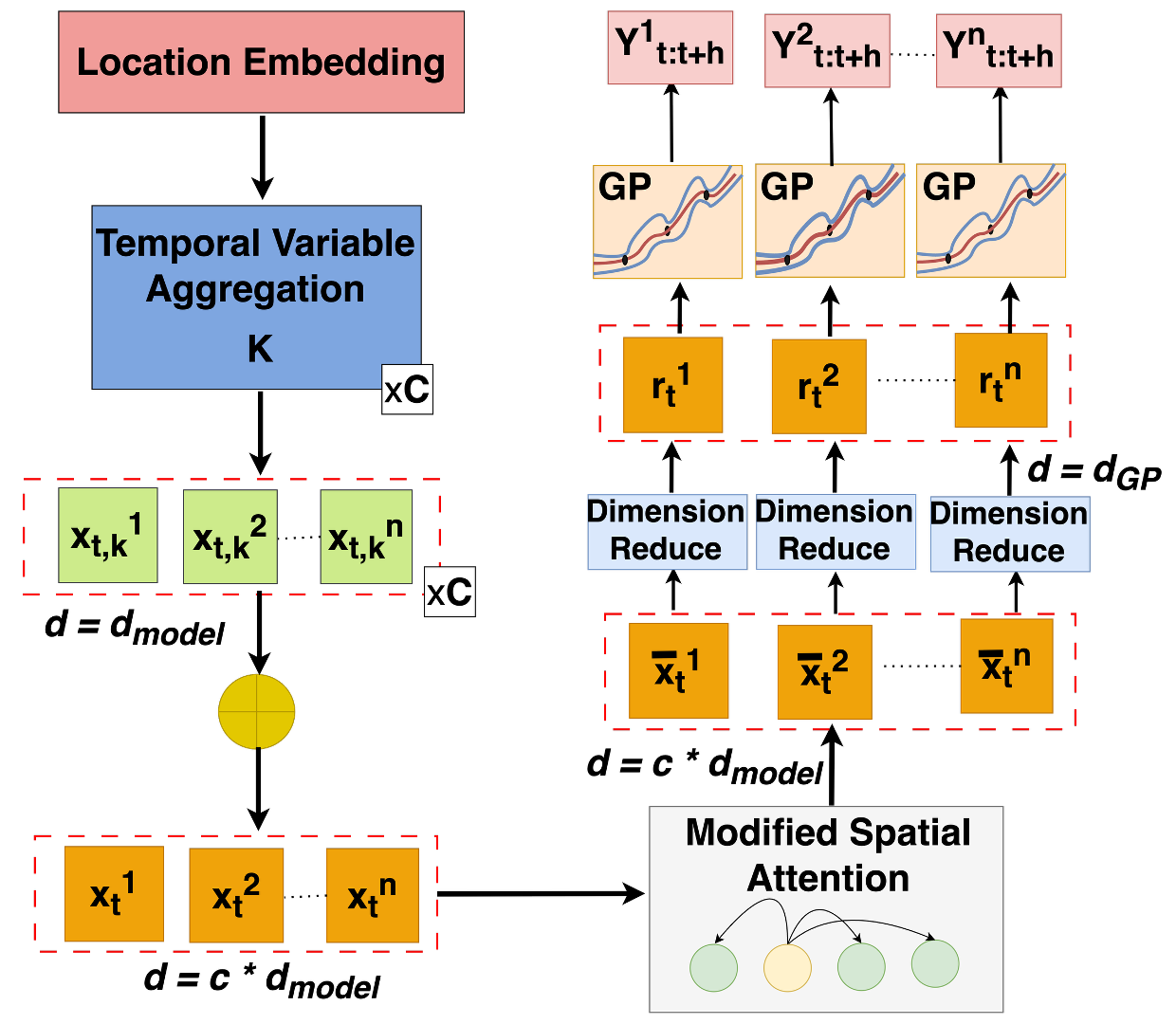}
            \centering
            \caption*{(a) Overall architecture}
        \end{minipage}
        % Right column with two stacked images
        \begin{minipage}{0.45\textwidth}
            \centering
            \includegraphics[scale=0.35]{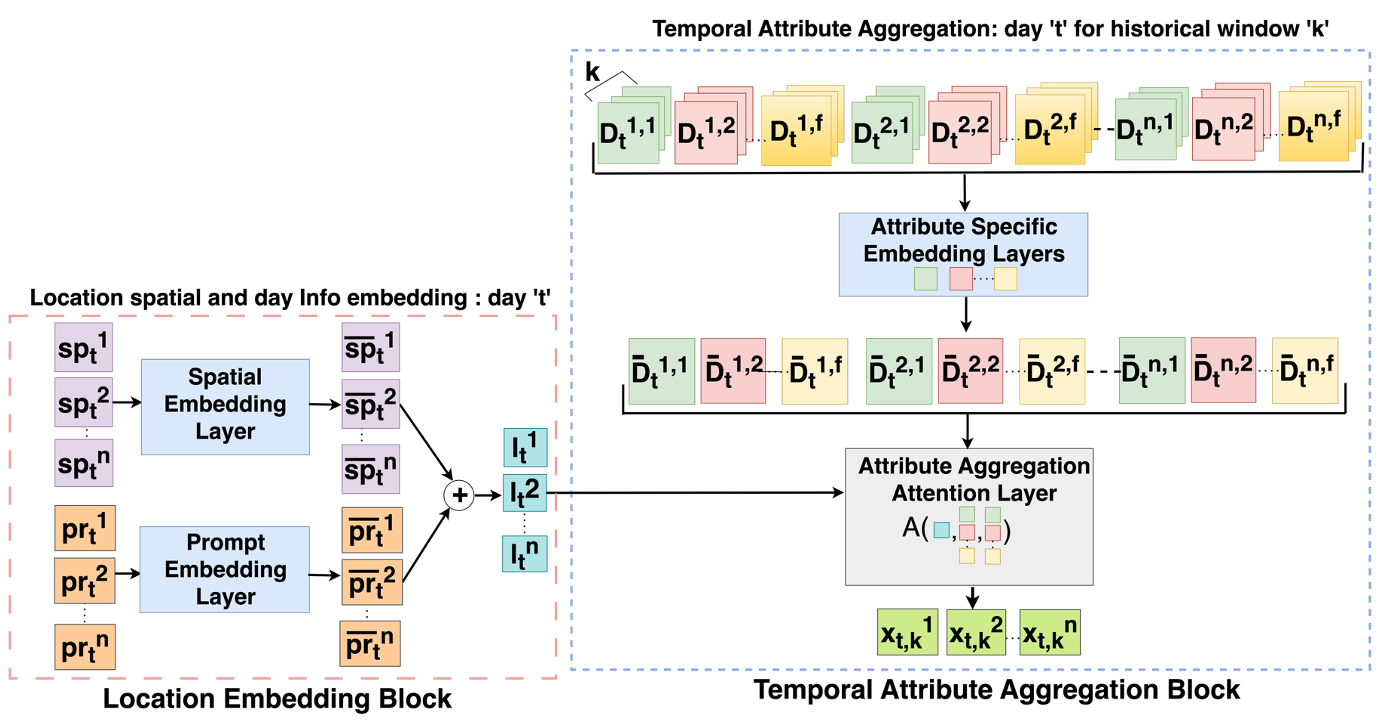} 
            \caption*{(b) Location embedding and attribute aggregation}
        \end{minipage}    
    \end{minipage}
\centering
\captionof{figure}{\fs{} model architecture (left) and its location embedding and temporal attribute aggregation components (right).   }
\label{fig:model_arch_overall}
\end{figure*}

% \delete{We pose the SWE forecasting problem as a spatio-temporal problem with a need for attribute aggregation, and formulate this as a two-step sequence modeling problem. Both steps involve implementing the concept of self-attention \citep{vaswani2017} to synthesize the necessary information for each token in the sequence. Finally, the Gaussian process \citep{Rasmussen06} forecasts SWE and provides its prediction interval, using the representation obtained from the spatial-temporal attention model.}

Our model design hinges on two core ideas. 
First, to incorporate spatio-temporal correlations as well as attribute interaction for SWE forecasting, we first model the problem as a sequence-to-sequence prediction problem, and present an adaptation of the self-attention mechanism \citep{vaswani2017}.
When used in real-time, this model will ingest the last $k$ days of data for each location and will output the SWE values for the next $h$ days (i.e., the forecast horizon), as shown in Figure~\ref{fig:intro_fig}.
Secondly, to implement a principled quantification of prediction uncertainty, we augment the model with Gaussian process (GP) \citep{Rasmussen06}, which operates on the distribution of data points rather than sequences. For any given point, GPs---with appropriate kernel selection and adequate input features---capture correlation across all the training data points to make predictions with uncertainty quantification. 
However, using GPs directly can introduce a limitation in the number of dimensions. Therefore, in our model architecture, we train our self-attention model to capture spatio-temporal correlation in the data at a lower dimensional space, followed by using GP as an output head to give forecast with a prediction interval.

Figure~\ref{fig:model_arch_overall}(a) shows the overall architecture of our model, and  Figure~\ref{fig:model_arch_overall}(b) shows the temporal attributes aggregation step of the model in detail.  

\paragraph{Notation:}$\mathbb{L}$ represents set of $n$ SNOTEL station; $\mathbb{S}$ is set of SWE seasons or water years (Oct. 1 to Sep. 30); $\mathbb{A}$ is a set of $f$ dynamic attributes attached to locations; $k$ is input historical window; $h$ is the forecasting horizon (in days or weeks); and $m$ is day number in a water year.

\textbf{Inputs}: Each training example is a sequence of all locations, their daily features and historical attributes, and prompts. The input sequences for day \textit{t} consist of:
\begin{itemize}[leftmargin=5pt]
%    \item $\textbf{pr}_t^{1}, \textbf{pr}_t^{2},...,\textbf{pr}_t^{n}$ represents a sequence of prompt vectors of $n$ locations denoted as $\textbf{Pr}_{t}$.
    
    \item{ $\textbf{Pr}_{t}$$=$$[\textbf{pr}_t^i]_{i=1}^n$:} a sequence of prompt vectors of $n$ locations.

    \item{$\textbf{Sp}_{t}$$=$$[\textbf{sp}_t^i]_{i=1}^n$:}  a sequence of spatial attribute vectors.

    % \delete{\item Each location $i \in \mathbb{L}$ has a collection of vectors of $f$ attributes with $k$ historical observations before day $t$, represented as $\mathcal{\textbf{D}}_{i,k}^{t}$ = $[\textbf{d}_{i,a}^{k,t}]_{a=1}^{f}$. For all $n$ locations, the sequence can be represented as 
    % $[\textbf{D}^{t}_{i,k}]_{i=1}^n$.}

    \item{ $\mathcal{\textbf{D}}_{t,k}[1$$:$$n,1$$:$$f,t-k$$:$$t]$$\in \mathbb{R}^{n \times f \times k}$:} a collection of $n$ locations with $k$ historical observations of $f$ attributes per  day $t$. For  location $i$, its attribute $a \in f $ is $\mathcal{\textbf{D}}_{t,k}^{i,a}$ = $\mathcal{\textbf{D}}_{t,k}[i,a,t-k:t]$. Here, $k$ can represent historical observations for different temporal resolutions as daily, weekly, monthly, and yearly.
\end{itemize}

% \delete{\textbf{Outputs}: The output of the model is a sequence of $n$ SWE forecast values  for the $n$ locations, along with prediction interval, for a forecast horizon of  $h$ days. For each location $i$ $\in$ $\mathbb{L}$ on day $t$, the forecast is a 2-D vector containing high ($u$), mean ($p$), and low ($l$) SWE respectively, given by $\textbf{Y}^{i}_{t,h}$ = $[[u_i^j]_{j=t}^{t+h}, [p_i^j]_{j=t}^{t+h}, [l_i^j]_{j=t}^{t+h}]$ which is referred to as the {\bf prediction interval}. Finally, for all $n$ locations on day $t$, the output sequence can be represented as $[\textbf{Y}_{t,h}^i]_{i=1}^{n}$.}

\textbf{Outputs}: The output of the model is a sequence of SWE forecast values for the $n$ locations, along with a prediction interval, for a forecast horizon of $h$ days. For each location $i$ $\in$ $\mathbb{L}$ on day $t$, the forecast is a 2-D vector containing high ($u$), mean ($p$), and low ($l$) SWE respectively, given by $\textbf{y}^i_{t:t+h} = [u^i_{t:t+h}, p^i_{t:t+h},l^i_{t:t+h}] \in \mathbb{R}^{3 \times h}$ which is referred to as the \textbf{prediction interval}. Finally, for all $n$ locations on day $t$, the output sequence can be represented as $\textbf{Y}_{t:t+h} \in \mathbb{R}^{n \times 3 \times h}$. $h$ can represent the forecasting horizon at different temporal resolutions such as daily and weekly points.

\paragraph{Location Embedding Block:} This block is divided into two parts as shown in the left part of Figure \ref{fig:model_arch_overall}(b).
    
  \underline{Spatial Embedding Layer} embeds \emph{daily spatial features}  
  (e.g., day number/length)  on day $t$ for each location, alongside key spatial features (e.g., lat/long, southness, elevation) to its corresponding embedding representation $\overline{\textbf{Sp}}_{t}$. For embedding we used $d_{model}$ = 1024 dimensions.  
 % While the spatial features are static, day information captures the location's behavior over time.
    
    \underline{Prompt Embedding Layer} provides prompts ($\overline{\textbf{Pr}}_t$) that represent additional context for the location  such as capture weather patterns, vegetation, and elevation range.

    These two embeddings provide the location-specific embedding for day $t$: $ \textbf{L}_t = \overline{\textbf{Sp}}_t + \overline{\textbf{Pr}}_t$. 

\paragraph{Temporal Attribute Aggregation Block:}
    % \delete{This is shown in the right part of Figure~\ref{fig:comp_arch}. The temporal attribute aggregation for each location is obtained individually. For any location $ i \in \mathbb{L}$ on the day $t$, given its $k$ historical observation of each attribute as $\mathcal{\textbf{D}}_{i,k}^{t}$, each attribute's historical vector is passed to their corresponding attribute embedding layers, finally to obtain $\mathcal{\overline{\textbf{D}}}_{i,k}^{t}$ = $[\overline{\textbf{d}}_{i,a}^{k,t}]_{a=1}^{f}$. 
    % %Each attribute is embedded in dimension $d_{model}$=1024.
    % Given location embedding vector $\boldsymbol{\ell}_{t}^{i} \in \textbf{L}_{t}$ and embedded attributes vectors $\mathcal{\overline{\textbf{D}}}_{i,k}^{t}$, we obtain temporal attributes aggregated representation as $\textbf{x}_{i}^{t}$---as explained in \S\ref{subsec:att_agg}.
    % Finally, for all $n$ locations, the sequence of aggregated attributes representation on the day $t$ for $k$ historical observations is obtained and denoted as $\textbf{X}_{t,k}$ = $[\textbf{x}_{i,k}^{t}]_{i=1}^{n}$.} %We use  $d_{model} = 1024$ for all the above embedding layers.}

     The temporal attribute aggregation for each location is obtained individually (Figure~\ref{fig:model_arch_overall}(b)). For  location $i$ on  day $t$, given its $k$ historical observation of each attribute as $\mathcal{\textbf{D}}_{t,k}^i = \mathcal{\textbf{D}}_{t,k}$ $[i,1:f,t-k:t]$, each attribute's historical vector is passed to their corresponding attribute embedding layers, finally to obtain $\mathcal{\overline{\textbf{D}}}_{t,k}^i$ = $\mathcal{\overline{\textbf{D}}}_{t,k}[i,1:f,1:d_{model}]$. 
    Given location embedding vector $\boldsymbol{\ell}_{t}^{i} \in \textbf{L}_{t}$ and its embedded attributes vectors $\mathcal{\overline{\textbf{D}}}_{t,k}^i$, we obtain the temporal attributes aggregated representation $\textbf{x}_{t,k}^i = \textbf{X}_{t,k}[i,1:d_{model}]$, where $\textbf{X}_{t,k}$ includes representation for all the locations.
        
% \delete{\paragraph{Window-wise Concatenation:} From the multiple temporal attributes aggregation blocks, we obtain $C$ different $\textbf{X}_{t,k_{c}}$, with their corresponding window $k_c$. These representations of locations are concatenated to obtain $\textbf{X}_t$, where the vector dimension for each location is $C \times d_{model}$.}

\paragraph{Window-wise Concatenation:} From the multiple temporal attributes aggregation blocks, we obtain $C$ different $\textbf{X}_{t,k_{c}}$, with their corresponding window $k_c$. These representations of locations are concatenated to obtain $\textbf{X}_{t}[1:n,1: C \times d_{model}]$, where the vector dimension for each location is $C \times d_{model}$. For  location $i$, its concatenated representation is obtained as $\textbf{x}_{t}^i = \textbf{X}_t[i,1:C \times d_{model}]$.
    
\paragraph{Modified Spatial Attention:} Next, to capture spatial correlations among locations, the attribute aggregated vectors for each location ($\textbf{X}_t$) are subjected to spatial attention ($\overline{\textbf{X}}_t$)---as explained in (\S~\nameref{sec:approach}). 
%This captures the spatial correlation between the locations. 
The dimension of each representation is  $C \times d_{model}$.

% \delete{\paragraph{Gaussian Process (GP):} Once all of the above blocks are trained using the actual SWE value, the spatio-temporal attention model is considered as the pre-trained model. To provide prediction uncertainty, we further replace its prediction head with a GP model \citep{Rasmussen06}. The GP prediction head can be learned using the dimensionally reduced ($d_{GP}(8) \ll d_{model}$) input representation of the pre-trained model, denoted by $\textbf{R}_{t}$, and its corresponding SWE output, as explained in \S\ref{subsec:gau_pr}.
% Note that for any location, when we pass its learned spatio-temporal representation $\textbf{r}_{t}^{i} \in \textbf{R}_{t}$ to the GP model, we obtain SWE forecast for horizon $h$ along with its prediction interval , represented by vector $\textbf{Y}^{i}_{t,h}$.}

\paragraph{Gaussian Process (GP):} Once all of the above blocks are trained using  actual SWE, the spatio-temporal attention model is considered as the pre-trained model. To provide prediction uncertainty, we  replace its prediction head with a GP model \citep{Rasmussen06}, which is learned using the dimension-reduced ($d_{GP}(8) \ll d_{model}$) input representation of the pre-trained model, denoted by $\textbf{R}_{t}$, and its corresponding SWE output.
%, as explained in (\S\nameref{sec:approach}).
Note that for any location, when we pass its learned spatio-temporal representation $\textbf{r}_{t}^{i} \in \textbf{R}_{t}$ to the GP model, we obtain SWE forecast for horizon $h$ along with its prediction interval, represented by vector $\textbf{y}^{i}_{t:t+h}$.

\section{Technical Approach}
\label{sec:approach}

In what follows, we describe the key components of our model architecture. 
%namely, attribute aggregation  to capture temporal correlations in the interactions between attributes, modified attention mechanism to capture global spatial correlation between locations, and multi-output Gaussian process to forecast SWE for each location along with prediction intervals.

\subsection{Attributes Aggregation}
\label{subsec:att_agg}
%Daily observed dynamic environment and meteorological attributes can have spatiotemporal variability, 
Attribute aggregation is used to capture  the attribute interactions  at a certain time of a location. Our approach inspired from \citep{nguyen2023climax} combines interactions of daily attributes into a vector representation.
%, capturing the historical temporal correlation of attributes at that location.
We implement the classical self-attention mechanism \citep{vaswani2017}, which works with a sequence of tokens  embedded into  query (\textbf{Q}), key (\textbf{K}) and value (\textbf{V}) vectors, as given by:
\begin{eqnarray}
\label{eqn:classic_att}
\hspace{-8mm}\mathrm{Attention}\big(\mathbf{Q},\mathbf{K},\mathbf{V}\big) &=& \rho\left(\frac{\mathbf{Q}.\mathbf{K^\top}}{\sqrt{d_\mathrm{model}}}\right)\mathbf{V} \ ,
\end{eqnarray}

where $\rho(.)$ represents the softmax function, and $d_\mathrm{model}$ represents the dimensions of all vectors.
%\ak{If space becomes an issue we can move the remainder of this subsection on attention to appendix, keeping only the main points as a summary.}
In our implementation, query
comes from daily spatial attributes and prompts for each location $i$  on day $t$, and is given by $\textbf{q}_{t}^{i}$ = $\textbf{sp}_{t}^{i}$ + $\textbf{pr}_{t}^{i}$. Meanwhile, key and value come from its $k$ historical collection of daily observations
($\mathcal{\textbf{D}}^{i}_{t,k}$). Therefore, the self-attention function 
%to aggregate historical attributes correlation within location $i$ $\in$ $\mathbb{L}$ in day $t$ $\in$ [1,m] of season $s$ $\in$ $\mathbb{S}$ and $f \in \mathbb{A}$ observed dynamic attributes 
is given by:

% \delete{\begin{eqnarray}\label{eqn:varAgg}
% \hspace{-5mm}\textbf{x}^{i}_{t,k} &=& \mathrm{Attention}\left(\textbf{q}_{t}^{i}.\textbf{W}_{tp}^{Q},\mathcal{\textbf{D}}_{i,k}^t.\textbf{W}_{tp}^{K},\mathcal{\textbf{D}}_{i,k}^t.\textbf{W}_{tp}^{V}\right) \ ,
% \end{eqnarray}}

\begin{eqnarray}\label{eqn:varAgg}
\hspace{-5mm}\textbf{x}^{i}_{t,k} &=& 
\mathrm{Attention}\left(\textbf{q}_{t}^{i}.\textbf{W}_{tp}^{Q},\mathcal{\textbf{D}}^{i}_{t,k}.\textbf{W}_{tp}^{K},\mathcal{\textbf{D}}^{i}_{t,k}.\textbf{W}_{tp}^{V}\right) \ 
\end{eqnarray}

%where $\textbf{x}^{i}_{t,k}$ $\in$ $\textbf{X}_{t,k}$ represents aggregated attributes representation for location $i$ on day $t$ in season $s$. 
$\textbf{W}_{tp}^{Q}$, $\textbf{W}_{tp}^{K}$ and $\textbf{W}_{tp}^{V}$ are weight matrices for query, key, and value respectively. Each attribute has historical information associated with it on day $t$, enabling the equation above to capture the temporal dynamics between attributes for each location.
%In addition to this, the next step focuses on implementing the spatial attention mechanism to capture spatial correlations between the locations. 
%This block operates on the dimension represented by $d_{model}$. 

\subsection{Modified Spatial Attention}
\label{subsec:mod_att}
%SWE values across the locations also are known to exhibit spatial correlation  \citep{thapa2024attention}. 
%Hence, capturing and exploiting all implicit spatial correlations between locations can improve SWE forecasting.
The location representation is the combination of representations of their attributes aggregated temporal dynamics from $C$ blocks, and is given by $\textbf{X}_{t}$ for all $n$ locations. It is then passed to the modified attention function $\overline{\mathrm{Attention}}$ to capture spatial correlations across the locations, 
\begin{eqnarray}\label{eqn:spAtt}
\overline{\textbf{X}}_{t} &=& \overline{\mathrm{Attention}}\Big(\textbf{X}_{t}.\textbf{W}^{Q}_{sp},\textbf{X}_{t}.\textbf{W}^{K}_{sp},\textbf{X}_{t}.\textbf{W}^{V}_{sp}\Big) \ 
\end{eqnarray}
Here, $\bar{\textbf{X}}_{t}$ is the encoded feature representation of all the $n$ locations in day $t$ $\in [1,m]$ of season $s$ $\in$ $\mathbb{S}$. $\textbf{W}_{sp}^{Q}$, $\textbf{W}_{sp}^{K}$ and $\textbf{W}_{sp}^{V}$ are the weight matrices. This encoding captures  spatiotemporal correlation information across and within locations. 

%The modified spatial attention function is an extension to the classical attention mechanism function (shown in Eqn.~\ref{eqn:classic_att}). 
The modified spatial function adds the Haversine distance \citep{Inman1835} and angularity between locations to derive the attention weights. The implementation details of Haversine distance and degree angularity can be found in Appendix. The rationale behind this is to add some bias of proximity to capture the spatial correlations between locations. However, the contributions of these additional elements in the attention weights are governed by learnable parameters, $\bm{w_{H}}$ and $\bm{w_{\theta}}$. Let $\textbf{Q}_{sp} = \textbf{X}_{t}.\textbf{W}_{sp}^{Q}$; $\textbf{K}_{sp} = \textbf{X}_{t}.\textbf{W}_{sp}^{K}$;  $\textbf{V}_{sp} = \textbf{X}_{t}.\textbf{W}_{sp}^{V}$ be the query, key and value vectors, respectively. The modified attention can be calculated as:
{\small 
\begin{eqnarray}\label{eqn:modAtt}
\hspace{-5mm}\overline{\mathrm{Attention}} \hspace{-3mm}&=&\hspace{-3mm} \rho\left( \rho\left(\frac{\textbf{Q}_{sp}.\textbf{K}_{sp}^{\top}}{\sqrt{d_{model}}}\right)
+ \bm{w_H} \cdot \textbf{d}_{H}  + 
\bm{w_{\theta}} \cdot \boldsymbol{\theta}\right)\textbf{V}_{sp} \ \end{eqnarray}
}

\begin{comment}
\begin{eqnarray}\label{eqn:modAtt}
 \overline{\mathrm{Attention}} &=&
 \rho\Bigg( \rho\!\left(\frac{\mathbf{Q}_{sp}\mathbf{K}_{sp}^{\top}}{\sqrt{d_{model}}}\right)
 + \mathbf{w_H}\cdot\mathbf{d}_{H} \nonumber \\
 &&\qquad\qquad + \mathbf{w_{\theta}}\cdot \boldsymbol{\theta} \Bigg)\mathbf{V}_{sp}
 \end{eqnarray}
 \end{comment}

\noindent where $\textbf{d}_H$ and $\boldsymbol{\theta}$ represent 2-D vectors for distance and angularity between all location pairs respectively. The variable aggregation and the spatial attention will collectively compute the encoded representation of spatiotemporal correlations across the locations for a given day $t$, as ($\overline{\textbf{X}}_t$). These encoded representations are dimensionally reduced to $d_{GP}$ $\ll$ $d_{model}$, transforming the sequence to $\textbf{R}_t$, and used to train Gaussian process~\citep{Rasmussen06} model to forecast SWE and quantify uncertainty. 

% For any two location pairs (a,b) $\in$ $\mathbb{L}$ on the day $t$, $f_{spatial}^{a,b}$ includes Haversine distance \citep{Inman1835}, $dist_{a,b}$ between location pairs and angularity $ang_{a,b}$ in addition to its attention weights. However, their influence is tuned with the learnable parameters $\bm{\alpha}_{dist}$ and $\bm{\beta}_{ang}$. Both distance and angularity are calculated using the latitude, longitude, and elevation of the locations. With all the available information, the modified attention can be calculated as,
% \begin{equation}\label{eqn:modAtt}
% \begin{split}
% f_{spatial}^{a,b} =  sm( sm(\frac{(\textbf{X}_{t}^{a,b}.\textbf{W}_{sp}^{Q})(\textbf{X}_{t}^{a,b}.\textbf{W}_{sp}^K)^{T}}{\sqrt{d_{model}}})\\ 
% + \bm{\alpha}_{dist} \times dist_{a,b}  + 
% \bm{\beta}_{ang} \times ang_{a,b}) (\textbf{X}^{a,b}_t .\textbf{W}_{sp}^V) 
% \end{split}
% \end{equation}

\subsection{Probabilistic Prediction with Gaussian Process}
\label{subsec:gau_pr}
%The forecasting of SWE will be uncertain because of various unforeseen circumstances. \KR{What do you mean by unforseen circumstances? I think you mean that the future (e.g. weather) is uncertain. What about the fact the model is just not perfect.}Hence, capturing the confidence interval of forecasted SWE for locations is important. 
To account for uncertainty in SWE forecasting, we replace the prediction head of the attention-based model in the previous sections with a Gaussian process (GP) regressor. This imposes a probabilistic prior on the prediction function $g(.)$ that maps from the pre-trained representation $\boldsymbol{r}_t^n$ at each spatio-temporal coordinate $(t, n)$ to its corresponding SWE value $y_{t:t+h}^n$ (see Figure~\ref{fig:gpimage}, Appendix).

To learn this prior, we remodel the pre-trained representation $\boldsymbol{r}_t^n$ as a tuple $\boldsymbol{z} = (\boldsymbol{r}, t)$ where $\boldsymbol{r} = \boldsymbol{r}_t^n$ and $t$ corresponds to its time index. This will allow the GP prior to model and learn the temporal correlation separately. We parameterize it as a $\tau$-component linear co-regionalized GP prior~\citep{van2020framework},
\begin{eqnarray}
\hspace{-4mm}g(\boldsymbol{z}) \hspace{-1mm}&\sim&\hspace{-1mm} \mathrm{GP}\Big(m(\boldsymbol{z}; \gamma), \kappa\big(\boldsymbol{z},\boldsymbol{z}'; \zeta_1, \zeta_2\big)\Big) \ \text{with}\ \nonumber\\
\hspace{-4mm}\kappa\big(\boldsymbol{z},\boldsymbol{z}'; \zeta_1, \zeta_2\big)
\hspace{-1mm}&\triangleq&\hspace{-1mm} \sum_{i=1}^\tau \zeta_2(t, i)\zeta_2(t',i) \cdot \kappa_i\big(\boldsymbol{r},\boldsymbol{r}'; \zeta_{1,i}\big)\ ,\label{eq:4.3.1}
\end{eqnarray}
where $\kappa_i(\boldsymbol{r}, \boldsymbol{r}'; \zeta_1)$ parameterizes the $i$-th component of the covariance between the overall spatio-temporal representation while the scalar product $\zeta_2(t, i)\zeta_2(t', i)$ parameterizes the $i$-th component of the correlation between time-steps $t$ and $t'$. Here, the kernel component $\{\kappa_i\}_i$ are parameterized as RBF kernels with learnable parameters $\zeta_1 = \{\zeta_{1,i}\}_{i=1}^\tau$ while $\zeta_2$ is a learnable matrix; and $m(\boldsymbol{z}; \gamma)$ is the mean SWE function parameterized by $\gamma$. Detailed specifications of these parameterization are provided in  Appendix.

\noindent {\bf Remark.}~Although temporal information is already embedded in the spatio-temporal representation $\boldsymbol{r}_t^n$, our formulation retains an additional, separate temporal correlation component. Without this, temporal correlation would be modeled implicitly through the overall spatio-temporal similarity, effectively imposing a single, possibly averaged time scale. However, in the SWE context, temporal dependencies can evolve differently over short and long time horizons \citep{nijssen2004effect}. By explicitly modeling temporal correlation, we allow for a more flexible structure that can capture the correlation across multiple temporal scales.

Let $\boldsymbol{y} = [g(\boldsymbol{z})]_{\boldsymbol{z}} = [g(\boldsymbol{r}_t^n, t)]_{(t,n)}$ denote the corresponding column vector of SWE ground-truth values. Let $\boldsymbol{m}_\gamma = [m(\boldsymbol{z}; \gamma)]_{\boldsymbol{z}} =  [m(\boldsymbol{r}_t^n, t; \gamma)]_{(t, n)}$ denote the column vector of the parameterized mean function's outputs at $\boldsymbol{z} = (\boldsymbol{r}_t^n, t)$ for all observed spatio-temporal coordinates $(t, n)$. Then,
\begin{eqnarray}
\hspace{-7mm}\boldsymbol{y} \hspace{-2mm}&\sim&\hspace{-2mm} \mathbb{N}\Big(\boldsymbol{m}_\gamma; \boldsymbol{K}_{\zeta}\Big) \ \text{where}\ \boldsymbol{K}_{\zeta}[\boldsymbol{z},\boldsymbol{z}'] \ =\ \kappa(\boldsymbol{z}, \boldsymbol{z}'; \zeta_1, \zeta_2) \ , \label{eq:4.3.2}
\end{eqnarray}
following the marginal property of GP. Having both the pre-trained input representation $\boldsymbol{z}$ and the corresponding SWE output $\boldsymbol{y}$, we can learn the parameters $(\gamma, \zeta = (\zeta_1, \zeta_2))$ via
\begin{eqnarray}
\hspace{-8mm}\big(\gamma, \zeta\big) &=& \argmax\ \log \mathbb{N}\Big(\boldsymbol{y}; \boldsymbol{m}_\gamma, \boldsymbol{K}_\zeta\Big) \ .
%\nonumber\\
%&=& \argmax \ -\frac{1}{2}\Big(\boldsymbol{y} - \boldsymbol{m}_\gamma\Big)^\top\boldsymbol{K}_\zeta^{-1}\Big(\boldsymbol{y} - \boldsymbol{m}_\gamma\Big) \ .\label{eq:4.3.3}
\end{eqnarray}
Given these GP parameters, the distribution over the true SWE value $y_\ast$ at any spatio-temporal coordinate $(t_\ast, n_\ast)$ with representation $\boldsymbol{z}_\ast = (\boldsymbol{r}_\ast, t_\ast)$ can be derived as the GP's predictive distribution~\citep{Rasmussen06},
\begin{eqnarray}
\hspace{-12mm}y_\ast &\sim& \mathbb{N}\Big(\boldsymbol{\kappa}_\ast^\top\boldsymbol{K}_\zeta^{-1}\Big(\boldsymbol{y} - \boldsymbol{m}_\gamma\Big) + m(\boldsymbol{z}_\ast; \gamma), \boldsymbol{\sigma}^2_\ast\Big) \ , \label{eq:4.3.4}
\end{eqnarray}
where  $\boldsymbol{\kappa}_\ast = [\kappa(\boldsymbol{z}_\ast, \boldsymbol{z}; \zeta_1, \zeta_2)]_{\boldsymbol{z}}$ denotes the column vector of covariance values between $\boldsymbol{z}_\ast$ and other observed spatio-temporal representations $\boldsymbol{z}$; and the predictive variance,
\begin{eqnarray}
\boldsymbol{\sigma}^2_\ast &=& \kappa\big(\boldsymbol{z}_\ast,\boldsymbol{z}_\ast; \zeta_1, \zeta_2\big) \ -\  \boldsymbol{\kappa}_\ast^\top\boldsymbol{K}^{-1}_\zeta\boldsymbol{\kappa}_\ast \ . \label{eq:4.3.5}
\end{eqnarray}
Eqs.~\eqref{eq:4.3.4} and~\eqref{eq:4.3.5} thus present an entire Gaussian distribution over our SWE prediction. This yields both the (mean) prediction,
\begin{eqnarray}
\hspace{-9mm}p_{n_\ast}^{t_\ast} &=& \mathbb{E}[y_\ast] \ =\ \boldsymbol{\kappa}_\ast^\top\boldsymbol{K}_\zeta^{-1}\Big(\boldsymbol{y} - \boldsymbol{m}_\gamma\Big) + m(\boldsymbol{z}_\ast; \gamma) \ , \label{eq:4.3.6}
\end{eqnarray}
and its prediction variance $\sigma_\ast^2$, which can be used to provably compute any $\alpha$-prediction interval $\boldsymbol{Y}_{t_\ast}^{n_\ast} = [\ell_{n_\ast}^{t_\ast}, u_{n_\ast}^{t_\ast}]$,
\begin{eqnarray}\ell_{n_\ast}^{t_\ast} &=& p_{n_\ast}^{t_\ast} \ -\ \Phi^{-1}(\alpha/2) \cdot \sigma_\ast \ ,\\
u_{n_\ast}^{t_\ast} &=& p_{n_\ast}^{t_\ast} \ +\ \Phi^{-1}(\alpha/2) \cdot \sigma_\ast \ , \label{eq:4.3.7}
\end{eqnarray}
where $\Phi^{-1}(.)$ is the inverse cumulative function of an univariate normal and $\alpha$ is the confidence level. For example, in our experiment, we choose $\alpha = 0.95$. This means under our learned GP calibration $(\gamma,\zeta)$, the true SWE value is in $[\ell_{n_\ast}^{t_\ast}, u_{n_\ast}^{t_\ast}]$ with $95\%$. The cost of the above training and inference procedure is however cubic in the number of spatio-temporal training data points $\boldsymbol{z} = (\boldsymbol{r}, t)$. To sidestep this prohibitively expensive cost, we can adopt existing sparse approximations~\citep{Candela05,Titsias09,van2020framework} of the above GP which scales the cost back to linear in the size of the training dataset.

\paragraph{Computational Complexity:} A detailed complexity analysis is provided in the Appendix. In a nutshell, the overall complexity of our model is governed by attribute aggregation, modified spatial attention, and Gaussian process blocks. %Here, the complexity is dominated by the number of inducing points in GP, locations in the modified spatial attention, and model dimension. 

\paragraph{Implementation:} 
Model implementation used Pytorch (v2.0.1) (LSTM and Attention models), GPyTorch (v1.12) (Gaussian process)
packages. Data processing and visualization used multiple
Python packages. This is an open-source project, and the code with data can be found in \textcolor{blue}{\url{https://github.com/Krishuthapa/SWE-Forecasting}}.

\section{Experimental Setup}
\label{sec:exp_setup}

\captionsetup{font={normalsize}}
\begin{figure*}[thb]
    \vspace{0pt}
    \centering
    \includegraphics[scale=0.60]{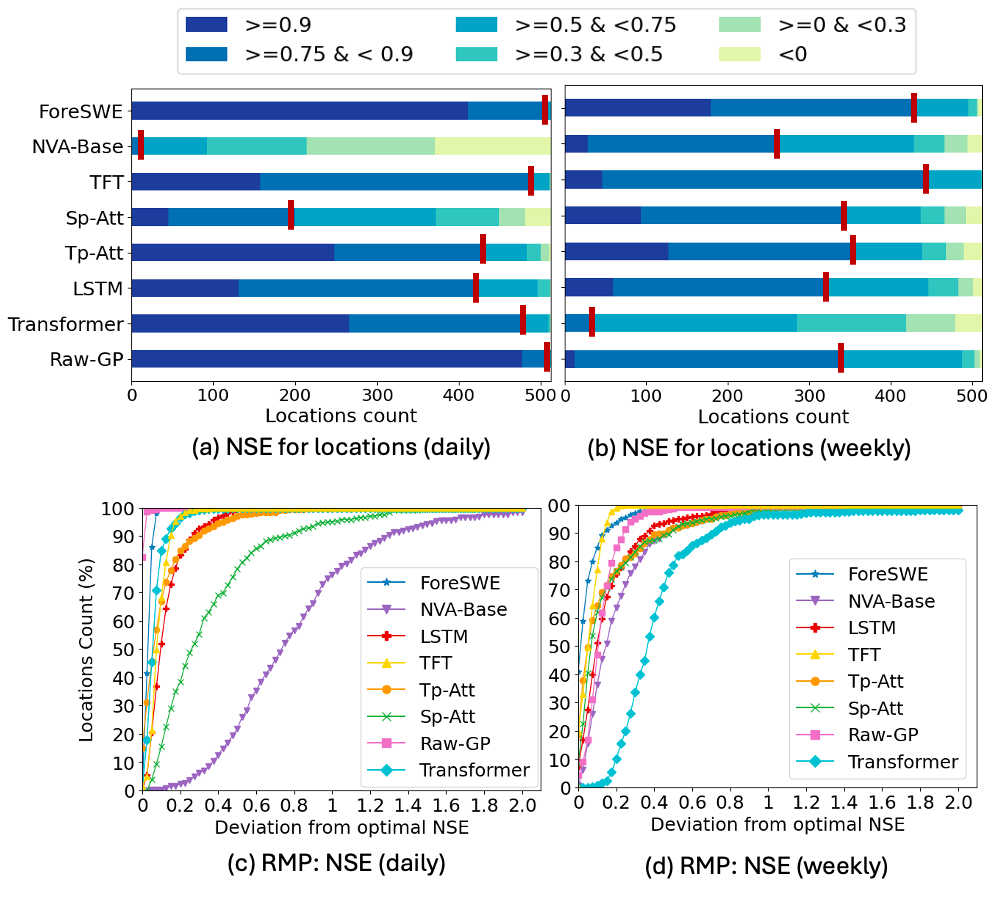}
    \caption{ {\bf Daily and Weekly forecasting} models comparison: (a) The distribution of locations across five NSE groups for all models. NSE is calculated for each location, with higher values (blue bars) indicating a better prediction. Locations to the left of the red line have NSE $>0.75$. (b) Relative model performance (RMP) based on the NSE values. RMP chart: Each curve corresponds to a model; the closer and longer the line along the y-axis, the better the model performance. The X-axis shows the deviation of a model from the corresponding best performing model; Y-axis shows the fraction of locations with that deviation.\vspace{-5mm}} 
    \label{fig:model_comparison}
\end{figure*}

 % \BS{Figure 4(a) seems like extended, also for Figure (b), on x-axis label use higher interval like an interval of 0.5 and for the y-axis also use an interval of 20 maybe. Just a suggestion if it makes the figure clean otherwise as it also works}
\paragraph{Data Description:}
\label{sec:Data}
The set of static and dynamic  features used along with their respective sources are listed below:
\begin{itemize}\itemsep=-0.05ex
   \item \textbf{Static Features}:
     elevation, latitude, longitude  \cite{NRCS:2023}; land cover \cite{Yang_2018}; southness 
     \cite{NED:2014}; prompt vectors generated from pre-trained language model \cite{reimers-2019-sentence-bert} based on vegetation, weather type, etc.  
    \item{\bf Daily Features:}
        SWE \citep{NRCS:2023}, max/min temperature, precipitation accumulation, downward surface shortwave radiation, wind velocity, max/min relative humidity, and specific humidity \citep{abatzoglou2013development}.
    \item{\bf Daily Satellite Observations:}
        Passive microwave brightness temperature (19VE, 37VE, and their difference) \citep{brodzik2016measures}
 \end{itemize}

 The daily \snotel{} data, downloaded from \cite{NRCS:2023}, consists of 822 stations for 28 water years (1991-2019).
We filtered out stations with more than 10\% missing snow observation data in any given year. The resulting 512 stations comprised our main data set. Given our focus on SWE forecasting, we used daily data for $\sim$180 days starting on Dec 1, to cover the active SWE season. This resulted in a total of 2,580,480 (=$512\times 28\times 180$) $\langle$location, year, day$\rangle$ combinations.
The satellite data for getting brightness reflectances, and Light Detection and Ranging (LiDAR) data for getting slope and elevation to calculate southness, following the approach discussed in \cite{thapa2024attention}.

%\KR{Does existing SWE forecast work you mention utilize satellite imagery? If we know the past SWE in a forecasting mode, it is unclear why past time series of satellite approximations of SWE is needed or how it may help. Except of course in areas outside of SNOTEL stations. Is that the intent or are you forecasting only for SNOTEL stations in this work? }
%\KT{we are only forecasting for SNOTEL stations. We are using brightness temperature data because we had used it in a previous paper. Also, also I vaguely remember losing accuracy when I tried to remove it.}
%\AK{Let's then keep the paragraph about satellite data source since you used it.}

\paragraph{Evaluation Methodology:}
\label{sec:evaluationmethodology}
Out of the 28 years, we used 25 years (from 1994) for training and testing
, splitting the 25 years into two sets: training (20 years), and testing (5 years). Here, we train and test our model on the same set of locations. The data from (1991-1993) is separated as a buffer to include yearly historical data (3 years) for inputs starting in 1994. Test water years are 2015 through 2019---consecutive so that the data points belonging to these years do not appear anywhere during our training. 
These test years also cover a wide range in average SWE, from driest (2015) to wettest (2017). 
%\KR{Is you evaluation not just for April through June timeframe or something like that? Is that specified in the paper. I missed it. So it unclear to me what figure 4 corresponds to.}
%\KT{ Its from Dec 1st to 180 days from that, somewhere until may 30. }
 Additionally, the locations were binned into four ``groups'' based on their averaged peak SWE, from lowest (group 1) to highest (group 4)---as  shown in Table~\ref{tab:locaton_groups} of Appendix.

 In our experiments, we compared eight models in a design ablation study (as outlined in Table~\ref{tab:model_param}, Appendix):
 \begin{itemize}\itemsep=-0.05ex
     \item{\lstm:} Long Short-Term Memory model in autoregressive encoder-decoder format \citep{hochreiter1997long}

     \item{\rgp:} Sparse Gaussian process with multi-variate output implemented on raw input features \citep{Rasmussen06}
    
    \item{\spatt{} and \tpatt:}  Spatial and temporal attention models, respectively, implemented using the attention mechanism presented in \citep{thapa2024attention}. These models are not autoregressive---i.e., they generate the next $h$ days of forecast from linear 
    transformation of their encoded representations.

    \item{\tar:} Temporal auto-regressive standard transformer \citep{vaswani2017}

    \item{\tft:} Temporal Fusion Transformer \citep{lim2021temporal}

    \item{\nva:} Non-Variable Aggregation is a trimmed down version of our proposed model without the variable aggregation and GP parts.

    \item{\bf \fs:} Our proposed uncertainty-aware attention model.

 \end{itemize}

 We evaluate models in two axes: forecasting accuracy and prediction interval. To compare models accuracy across location groups, we use Nash-Sutcliffe Efficiency (NSE) \citep{nash1970river}, which compares the predicted ($Y_p$) versus actual ($Y_a$) over the entire prediction time ($T$) for any location $i\in\mathbb{L}$.
 Its value ranges from -$\infty$ to 1, where the model performs best with its value closer to 1. Generally, in a long-horizon forecasting setup, \textbf{NSE $>$ 0.75} is preferred \citep{Moriasi2007}. A value below 0 means an averaged extrapolation would be better than the model prediction. 

\begin{equation}\label{eqn:NSEformula}
 NSE_{i}  =  1- \frac{\sum_{t=1}^{T} (Y_{p,i}^{t} - Y_{a,i}^{t})^{2}}{\sum_{t=1}^{T} (Y_{a,i}^{t} - \overline{Y}_{a,i}^t)^2}
\end{equation}
Here,  ${\overline{Y}}_{a,i}$ is the long-term mean actual SWE for location $i$.  We also use relative bias that quantifies over-/under-predictions: 
%, a metric to obtain the percentage of under- and over-prediction by the models:
 %For any location $i\in$ $\mathbb{L}$, its relative bias is given by:

  {\small
  \begin{equation}\label{eqn:RBformula}
 RB_{i}  =  \frac{\sum_{t=1}^{T} (Y_{p,i}^{t} - Y_{a,i}^{t})}{\sum_{t=1}^{T} Y_{a,i}^{t}}   
 \end{equation}
 }

%where, $T$ is the number of distinct year-month-days used in testing,  $Y^t_{a,i}$ and $Y^t_{p,i}$ are actual and forecasted SWE respectively on a given day $t\in[1,T]$, and  ${\overline{Y}}_{a,i}$ is the long-term mean actual SWE. 

Additionally, to evaluate predictions with uncertainty quantification from \fs~and \rgp, we use metrics such as Negative Log Likelihood (NLL), Expected Calibration Error (ECE) and coverage percentage, further explained in the Appendix.
%~\ref{sec:gaussian_process_metrics}. 
%The best model has a high coverage percentage with low NLL and ECE.
%The positive value shows the overprediction percentage and vice versa, with the ideal bias at 0\%.

As for the \textit{forecasting horizon}, we trained and tested our model under two experimental settings: forecasting daily SWE for a 10-day horizon, and forecasting weekly SWE for 4-week horizon.

\section{Experimental Results}
\label{sec:results}
This section presents and analyzes detailed evaluation of our approach in terms of performance and uncertainty quantification in the daily and weekly forecasting setups.

%We trained our models to forecast daily SWE for 10 days and weekly SWE for 4 weeks. The results below are based on the 10-day forecasts. The weekly forecasting results are similar, as shown in the Appendix. 
%~\ref{sec:appendix_plots}.

\captionsetup{font={normalsize}}
\begin{figure*}[tb]
    \centering
    \includegraphics[scale=0.80]{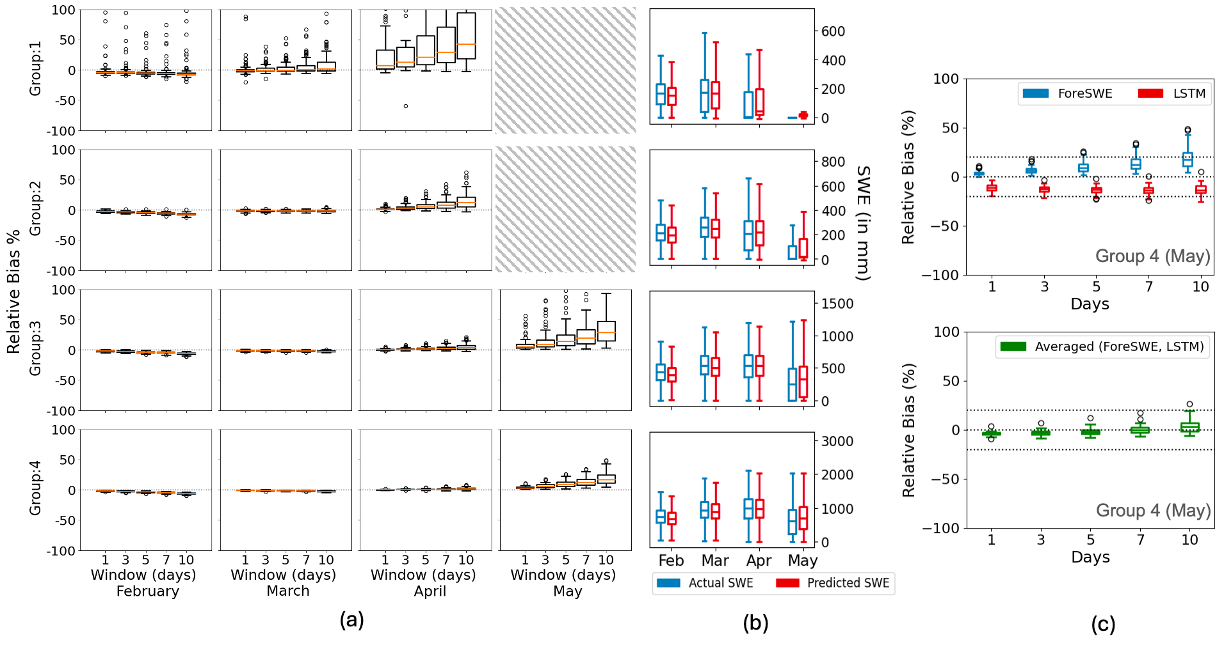}\vspace{-4mm}
    \caption{\normalsize Daily Forecasting: (a) \fs~ model's  relative bias, by the different months and location groups (with forecasting horizons ranging from 1 to 10 days). Groups 1 and 2 are blocked for May as the snow has melted completely at these locations. (b) Actual and \fs{} Predicted SWE availability in different groups across the active SWE months. (c) (upper) Relative bias of \fs~ model against a temporal model (\lstm) for Group 4 in May over different forecasting windows. (lower) Relative bias with the ensembling of \fs~ and \lstm~ for Group 4 in May. The dotted lines in the plot mark $\pm$ 20\%.} 
    \label{fig:daily_fgp_months}
\end{figure*}

\paragraph{Model Comparison:} 
% \delete{We evaluate the model's performance using NSE values and relative bias (\%). For daily forecasting, Figure 4(a) shows that our proposed model, \fs{}~outperforms all other models in terms of the fraction of locations with high NSE values---in particular, 99.2\% of the locations have NSE $>0.75$ with \fs{}, while the next best model is \tpatt with 80\% of the locations. Figure 4(b) shows the relative performance chart which shows that \fs{} achieves the best NSE value for nearly 90\% of the locations, and when it is not the best, it is still within 0.1 of the best NSE for the remaining locations.
% A similar trend is observed for relative bias in Figure 4(c), \fs{} has the least relative bias for 90\% of locations. 
% Among the other models, \tpatt{} and \lstm{} perform reasonably well.
% %For weekly forecasting, as shown in Figure 8 (Appendix \ref{sec:appendix_plots}), \fs~(83.2\%) outperforms next best \tpatt~(68.7\%) in terms of proportions of locations with NSE $>$ 0.75. 
% While weekly forecasts show a similar trend (Figure~\ref{fig:model_comparison_weekly}; Appendix \ref{sec:appendix_plots}), there is a marginal drop in forecasting accuracy across all models (to be expected with the longer horizon). }

We evaluate the model's performance using NSE values and relative bias (\%).  Figure~\ref{fig:model_comparison} shows model comparison based on NSE for daily and weekly forecasting. 

For \textit{daily} forecasting, we observed that \fs{} and \rgp{} models achieve the best NSE values with  comparable performance. In particular, \fs{} and \rgp{} achieve high NSE ($>0.75$) in 99.2\% and 99.6\% of the locations, respectively (Figure~\ref{fig:model_comparison}a). 
%Both \rgp{} and \fs{} significantly outperform other sequence-based models, where the next best model is \tar{} with 93.7\%. 
%%%%%%Since \fs{} and \rgp{} forecast SWE with a prediction interval, for this set of results, we are using the mean prediction. 
Figure~\ref{fig:model_comparison}c shows the relative performance chart based on the NSE values comparing the different models against one another. Notably, temporal models perform better, indicating the prominence of temporal effect for near-term forecasting. 
%Both \rgp{} and \fs{} achieve the best NSE values for 90\% of the locations, clearly outperforming all other models.

For \textit{weekly} forecasting,  \fs{} achieves the best performance. Figure~\ref{fig:model_comparison}b shows that \fs{} achieves NSE above 0.75 at over 427 locations while \rgp{} only achieves comparable NSE at only 340 locations. This performance advantage of \fs{} over \rgp{} is also reflected in the relative model performance chart (Figure~\ref{fig:model_comparison}d). This shows that \fs{} is the preferred model for the longer forecasting horizon (i.e., weekly), which makes it particularly useful in sub-seasonal scale decision making. Interestingly, the relative performance of \tpatt{} is comparative or outperforms \rgp{} for the weekly horizon (Figure~\ref{fig:model_comparison}d).  
Furthermore, in weekly forecasting \spatt{} outperforms most of the temporal models such as \lstm{} and \tar{}, underscoring the significance of spatial correlation in relatively long-term forecasting.

%\ak{We need a good justification here to discuss \rgp{} and \fs{} performances. Currently it is only explaining the observation from the result but does not explain why we are seeing what we are seeing. @Nghia/Krishu - can you please help with this?}
\begin{table}[htbp]
\vspace{0pt}
\centering
\small
\setlength\tabcolsep{4pt}  % reduce column padding
\begin{tabular}{llccccc}
\toprule
\textbf{Metric} & \textbf{Model} & \textbf{2015} & \textbf{2016} & \textbf{2017} & \textbf{2018} & \textbf{2019} \\
\midrule
\multirow{2}{*}{NLL} & \fs & \textbf{6.94} & \textbf{7.48} & \textbf{7.48} & \textbf{6.66} & \textbf{6.49} \\
                     & \rgp  & 11.97 & 16.85 & 22.70 & 17.46 & 18.42 \\
\midrule
\multirow{2}{*}{ECE} & \fs & \textbf{0.14} & \textbf{0.15} & \textbf{0.18} & \textbf{0.12} & \textbf{0.15} \\
                     & \rgp  & 0.362 & 0.38 & 0.41 & 0.37 & 0.40 \\
\midrule
\multirow{2}{*}{Coverage} & \fs & \textbf{80.88} & \textbf{79.57} & \textbf{76.12} & \textbf{82.53} & \textbf{79.84} \\
                          & \rgp  & 58.71 & 56.62 & 53.89 & 57.15 & 54.54 \\
\bottomrule
\end{tabular}
\caption{Reported qualities of uncertainty estimates produced by \fs{} and \rgp{} over the years (detailed in Table~\ref{tab:daily_gp_metric_std}, Appendix). The reported metrics include NLL (Negative Log Likelihood), ECE (Expected Calibration Error), and Coverage.~Higher coverage and lower values for NLL and ECE indicate better uncertainty estimates. }
\label{tab:daily_gp_metric}
\end{table}

Furthermore, as observed from Fig.~\ref{fig:model_comparison}c and Fig.~\ref{fig:model_comparison}d, the performance of \rgp{} decreases substantially in the weekly forecasting setup while \fs{} still maintains its top performance.~This reveals a weakness of \rgp{} in spatio-temporal modeling.~It operates in the raw feature space and over-smooths itself to the short-scale variation in observed data, over-emphasizing temporal correlations over spatial correlations. This helps \rgp{} generalize well in short-term daily forecasting but mislead it in long-term weekly forecasting where spatial correlation has a stronger impact on the snow pattern. In contrast, \fs{} operates in the embedding space of a spatio-temporal transformer which were pre-trained to generate the most holistic spatio-temporal representation of data. This allows \fs{} to preserve its best performance in both short-term and long-term forecasting, underscoring the importance of using deep spatio-temporal representation for both short-term and long-term snow forecasting.

%In a short-term forecasting setup, temporal correlation is more prominent (as seen in Figure~\ref{fig:model_comparison}(c)) and nearby locations mostly have similar behavior; \rgp{} operates in the raw feature space, and the kernel function captures temporal correlation, while minimizing possible noise from other locations' data points. However, in long-term forecasting, temporal correlation itself is not enough, and spatial correlation between farther locations can also have significance(Figure~\ref{fig:model_comparison}(d)). Here, \fs{} benefits from the spatio-temporal representation of data points by bringing similar points closer, irrespective of their spatial and temporal proximity.\kt{Professor Nghia -- Can you please verify this edit?} 

\paragraph{Uncertainty in Forecasting:} 
In addition to predictive performance, prediction uncertainty is also important for downstream decision making. To evaluate this aspect, we further evaluate and compare the prediction uncertainty of \fs{}  and \rgp{} over entire intervals of the test years. The models were compared using the NLL, ECE and coverage metrics (defined in Appendix) where it is desirable to have a high coverage percentage with low NLL and ECE values.
%Metrics to evaluate the uncertainty in prediction include NLL, ECE and coverage percentage. 
%A lower value of NLL and ECE is expected, which shows that the model captures correct outcomes with high probability and has better confidence interval calibration, respectively. Similarly, a better-performing model is expected to have a higher coverage percentage. 
In the \textit{daily} forecasting setup, Table~\ref{tab:daily_gp_metric} shows that \fs{} has better uncertainty estimates than \rgp{} across all metrics for all the test years, indicating that it is a more suitable tool to be integrated into downstream decision making processes. 
We also have similar observations in the \textit{weekly} forecasting setting (see Table~\ref{tab:weekly_gp_metric} in Appendix). Both methods also have lower coverage in the weekly setting compared to the daily setting which is expected since forecasting for longer horizons is generally associated with higher uncertainty.~An example output for \fs{} is shown in Figure~\ref{fig:gpimage} of Appendix. 

To further understand how the \fs{} forecast skill varies by the snow period, we analyzed the locations classified into four groups based on average SWE (group 1: low, to group 4: high;  Table~\ref{tab:locaton_groups}; Appendix).  
Figure~\ref{fig:daily_fgp_months}(a) shows how \fs~ model performs across the different months, locations, and forecasting horizons, under the daily forecasting setting.  Figure~\ref{fig:weekly_fgp_months} in Appendix shows for weekly forecasting. Our key observations are as follows:
\begin{itemize}[leftmargin=5pt]
\itemsep=-0.05ex
\item Forecasting accuracy is high in the active snow accumulation phase (Feb, Mar) across all location groups. March signifies  an onset of the melting phase in low SWE locations (group 1), which accounts for a slight increase in error.
\item 
April is when most locations reach their peak SWE with an accelerated melt phase and the forecast errors are higher for groups 1 and 2 (Figure~\ref{fig:daily_fgp_months}(b)). 
However, groups 3 and 4, which have larger snowpacks, have a median relative bias close to 0\%. 
\item
In May, groups 3 and 4 show increasing  relative bias with horizon, as most of the snow is melted (Figure~\ref{fig:daily_fgp_months}(a)). %We don't consider groups 1 and 2 here, because of their almost complete snowmelt.
Furthermore, we observed that a fully-temporal model such as \lstm{} consistently underpredicts in May, while our \fs{} model overpredicts (Figure~\ref{fig:daily_fgp_months}(c))---suggesting the suitability of an average ensemble of these two models during this low-snowpack month.
\end{itemize}

\vspace*{-0.03in}
\section{Road to Deployment}
\label{sec:road_to_deployment}

In this paper, we presented and demonstrated the effectiveness of a new uncertainty-aware attention model to forecast SWE in real-world settings. Our approach combines the strengths of the attention mechanism in exploiting spatial and temporal correlations, with the strengths of GP in uncertainty quantification.

To support deployment of this new model, our team is actively pursuing collaborative directions, building on existing partnerships with regional and state water management agencies. For example, we have engaged with a regional office of the U.S. Bureau of Reclamation (USBR) to outline the design of a dashboard that can  inform their reservoir operations—a tool that is currently lacking. We plan to co-develop this dashboard with them to ensure it meets operational needs.
Additionally, multiple regional agencies involved in water management coordinate through the regional River Forecast Center. Once a dashboard is implemented, we intend to broaden our engagement to include these additional partners.

Through ongoing interactions with stakeholders, two critical requirements for deploying AI models have emerged: i) prediction explainability at the individual forecast level, by integrating interpretability tools (e.g., SHapley Additive exPlanations) and ablation studies, to provide actionable explanations for each prediction; and ii) transparent uncertainty quantification, which we have prioritized in the current work to bring us closer to operational readiness. 

SWE is an important intermediate variable for streamflow forecasting, and with the uncertainty-aware SWE forecasting over different horizons, \fs{} is capable of providing a well-calibrated input in real-time into streamflow forecasting models that USBR and its contemporaries internally uses for their reservoir management decision workflows. 

\paragraph{\bf Future directions:}
In addition to deployment efforts, we also plan to continue improving and extending the \fs{} model---including expanding its applicability to previously unseen locations and incorporating feedback from end users as deployment progresses, enabling more robust and user-informed modeling. Feature-based ablation studies can further improve the interpretability of the model. 

\begin {comment}
To deploy this new model, our team plans to pursue two major collaborative directions, leveraging already existing partnerships with the related regional and state agencies. 
First, our team plans to work closely with our partners in NRCS  to integrate our model into the NRCS SWE Dashboard \cite{usda_nrcs_snow_water_interactive_map}. The current dashboard does \emph{not} provide any short-term or subseasonal forecasting function, nor has a way to incorporate uncertainty. With a demonstrated ability to forecast for multiple horizons and with uncertainty bounds, our \fs{} model is well-positioned to complement the current capabilities of this dashboard, which is used nationwide for real-time SWE monitoring.

Secondly, our team plans to work closely with USBR and its contemporaries like Northwest River Forecast Center (NWRFC), US Army Corps of Engineers (USACE) in providing the SWE forecasts that would enhance decisions making related to reservoir management. In particular, SWE is an important intermediate variable for streamflow forecasting, and with the uncertainty-aware SWE forecasting over different horizons, \fs{} is capable of providing a well-calibrated input in real-time into streamflow forecasting models that USBR and its contemporaries internally uses for their reservoir management decision workflows. 

In addition to the above planned deployment-driven projects, we also plan to improve and extend our \fs{} model including by (a) improving interpretability of the model through ablation studies; (b) extension to unseen locations; and (c) improved modeling through community feedback that results from deployment. 

\end{comment}

\section{Acknowledgements}
\label{sec:ack}
This research was supported by USDA NIFA award No. 2021-67021-35344 (AgAID AI Institute).

\bibliography{aaai2026}

\clearpage

\appendix
\section{Appendix: Equations of Added Implementation In Modified Spatial Attention}
\label{sec:appendix_mod_sp}

This section discusses different mathematical techniques implemented while framing solutions for forecasting tasks in the paper.

\subsection{Haversine Distance}
\label{subsec:haversine}

Haversine distance captures the angular distance between two points on the surface of a sphere. In the paper, we used to find the distance between locations using their latitude and longitude, which was then used to include learnable proximity bias in capturing spatial attention between locations. Given two locations $a,b \in L$ with respective latitudes ($\boldsymbol{\varphi_a}$, $\boldsymbol{\varphi_b}$) and longitudes $(\boldsymbol{\phi_a},\boldsymbol{\phi_b})$, Haversine distance ($d_h$) can be calculated as,

\begin{equation}
\small
    d_H = 2r.arcsin\Bigg(\sqrt{sin^2\big(\frac{\boldsymbol{\Delta}_{\varphi}}{2}\big) + cos(\boldsymbol{\varphi_a}).cos(\boldsymbol{\varphi_{b}}).sin^2\big(\frac{\boldsymbol{\Delta}_{\phi}}{2}\big)}\Bigg)
\end{equation}

\noindent where, r is Earth's radius, and $\boldsymbol{\Delta}_{\varphi}$ and $\boldsymbol{\Delta}_{\phi}$ are difference in latitude and longitude, respectively. 

\subsection{Angularity Between Locations}
\label{subsec:angularity}

The degree of angularity between two locations can obtained by transforming their latitude, longitude, and elevation to 3D Cartesian coordinates. Given two locations $a,b \in L$ with respective latitudes ($\boldsymbol{\varphi_a}$, $\boldsymbol{\varphi_b}$), longitudes $(\boldsymbol{\phi_a},\boldsymbol{\phi_b})$ and $elevations(\boldsymbol{\epsilon}_{a}, \boldsymbol{\epsilon}_{b} )$, angularity ($\boldsymbol{\theta}$) can be calculated in following steps,

\begin{enumerate}
    \item Transforming to 3D Cartesian coordinates, given latitude ($\boldsymbol{\varphi}$), longitude ($\boldsymbol{\phi}$) and elevation ($\boldsymbol{\epsilon}$),as 
    \[ x = (r + \boldsymbol{\epsilon}).cos(\boldsymbol{\varphi}).cos(\boldsymbol{\phi})\]
    \[ y = (r + \boldsymbol{\epsilon}).cos(\boldsymbol{\varphi}).sin(\boldsymbol{\phi})\]
    \[ z = (r + \boldsymbol{\epsilon}).sin(\boldsymbol{\varphi})\]

    here, $r$ is Earth's radius.
    
    \item Calculate cosine similarity, for locations coordinates, $\textbf{c}_{a}$ = $[x_a,y_a,z_a]$ and $\textbf{c}_{b}$ = $[x_b,y_b,z_b]$ as,
        \[cos(rad) = \frac{c_a.c_b}{|c_a||c_b|}\]

    \item Finally angle in degrees $(\boldsymbol{\theta})$ can be obtained as,
    \[ Angularity(\boldsymbol{\theta}) = \frac{180}{\pi} \times rad\]
\end{enumerate}

\section{Appendix: Kernel Parametrization for Gaussian Process}
\label{sec:appendix_gp}

The overall kernel in our GP implementation is sum over $\tau$ components; each component $i$ combines spatial and temporal information, given by:
\begin{eqnarray}
    \kappa(\boldsymbol{z},\boldsymbol{z}'; \zeta_1,\zeta_2) = \sum_{i}^{\tau} \zeta_2(t,i).\zeta_2(t',i).\kappa_i(r,r';\zeta_{1,i})
\end{eqnarray}

$\zeta_2(t, i)$ is a learnable \textbf{temporal mixing weight}, which shows how much the latent component $i$ contributes at time $t$, and is represented as,
\begin{eqnarray}
    \zeta_2(t,i) = B_i^{\intercal}.h_t
\end{eqnarray}

where, $B_i$ is a learnable matrix for component $i$ and $h_t$ is the temporal feature at time $t$.

The \textbf{radial basis function (RBF)} kernel models the covariance between two inputs $\textbf{r}$,$\textbf{r}'$ based on an exponentially decaying function of their distance, which is parameterized by a signal and length-scale parameter. Following our notation in the main text, the parameterization of the $i$-th RBF component in our $\tau$-component co-regionalized GP prior is parameterized with $\zeta_{1,i} = \{\ell_i, \sigma_i\}$ as detailed below,
\begin{eqnarray}
\kappa_i\big(\textbf{r},\textbf{r}';\zeta_{1,i}\big) &=& \sigma^2_i \exp\left(-\frac{||\mathbf{r} - \mathbf{r}'||^2}{2\ell^2_i}\right)  \ ,
\end{eqnarray}
where the length-scale $\ell_i$ the smoothness of the GP-distributed function while the signal $\sigma^2_i$ controls its amplitude. Intuitively, a random function sampled from the corresponding GP prior will vary slowly with a large value of $\ell_i$ and vice-versa. On the other hand, a large value for the signal parameter $\sigma_i$ will influence the function to have larger variations and vice versa.

\section{Appendix: Metrics to Evaluate Probabilistic Outputs from Gaussian Process in \fs{} and \rgp{}}
\label{sec:gaussian_process_metrics}

\subsection{Negative Log Likelihood}
\label{subsec:neg_ll}

Negative Log Likelihood (NLL) measures the likelihood of actual values, assuming they were drawn from the predicted values distribution. A lower value of NLL means the model mean prediction is close to the true value, and the prediction interval is well-calibrated. Mathematically, it can be represented as,

\begin{eqnarray}
\boldsymbol{NLL(y;\mu,\sigma^2)} = \frac{1}{2} log(2\pi \sigma^2) + \frac{(y - \mu)^2}{2\sigma^2}
\end{eqnarray}

where, $y$ is true value; $\mu$ and $\sigma$ represent predicted mean and variance, repsectively. The first term $\frac{1}{2} log(2\pi \sigma^2)$ penalizes the over-confidence in the prediction and the second term $\frac{(y - \mu)^2}{2\sigma^2}$ penalizes inaccuracy.

\subsection{Expected Calibration Error}
\label{subsec:ece_gp}

Expected Calibration Error (ECE) evaluates if the model's predicted probabilities are trustworthy; it measures the average absolute difference between the expected coverage and empirical coverage. It is an essential metric when decisions rely on uncertainty. Mathematically, it can be represented as,

\begin{eqnarray}
\boldsymbol{ECE} = |coverage - confidence\_interval|
\end{eqnarray}

where coverage holds the fraction of actual values that fall inside the model's prediction interval.

\section{Appendix: Computational Complexity}
\label{sec:computation_complexity}

The computational complexity of the model is governed by three architecture design choices:
\begin{itemize}
    \item \textbf{Temporal Attribute Aggregation:} This block utilizes an attention mechanism (eqn~\ref{eqn:varAgg}) to capture temporal attributes aggregation for all locations on any day $t$. In the implementation, query represents spatial features of locations i.e. $\mathbb{R}^{n \times 1 \times d_{model}}$; key and value represent a collection of $f$ attributes with $k$ historical observations for all the locations, i.e. $\mathbb{R}^{n \times f \times d_{model}}$. The space and time complexities are given by $O(n.f.d_{model} + n.f)$ and $O(n.f.d_{model})$, respectively. Here, the impact of selecting a longer historical window $k$ is minimal, as historical observations for each attribute are embedded in $d_{model}$ regardless of its length.

    \item \textbf{Modified Spatial Attention:} This block performs a spatial attention mechanism (eqn~\ref{eqn:modAtt}) on the interaction of aggregated attributes, concatenated over $C$ historical windows for $n$ locations on any day $t$, i.e. query, key, and value are in $\mathbb{R}^{1 \times n \times (C \times d_{model})}$. Therefore, the space and time complexities are given by $O(n.C.d_{model} + n^2)$ and $O(n^2.C.d_{model})$, respectively. The resource requirement increases quadratically with location count; forecasting for a sequence with a large number of locations becomes expensive. A large model dimension also affects the complexity of the model.

    \item \textbf{Gaussian Process:} Each input in the Gaussian process (eqn~\ref{eq:4.3.2}) is a combination of (location, season, day), making a total of $ N = n \times s \times m$ dimensionally reduced ($d_{GP} = 8$) representations. The total datapoints are in the order of $\approx10^6$, hence we use a sparse Gaussian process with multivariate output for forecasting, with $p \ll N$ inducing points and $l$ latent GPs shared across $h$ forecasts. Therefore, space and time complexities are given by, $O(l.p.d_{GP} + l.p^2+ N.l.p)$ and $O(l.N.p^2 + l.p^3 + N.p.l.d_{GP} + h.l)$, respectively. A larger set of inducing points and latent function count imposes a significant computational burden.
\end{itemize}

\section{Appendix: Code And Data}
\label{sec:code_data_appendix}

The code and data for this work can be found in: \textcolor{blue}{\url{https://anonymous.4open.science/r/SWE-Forecasting-C723}}

\onecolumn

\section{Appendix: Additional Results}
\label{sec:appendix_plots}

\begin{table*}[tbh]
\centering
\begin{tabular}{|m{1.2cm}|m{2.1cm}|m{1.2cm}|m{2.3cm}|}
%\begin{tabular}{|c|c|c|c|}
\hline
\textbf{Group} & \centering \textbf{Average Max SWE(mm)} & \centering \textbf{Location \newline Count} & \textbf{Elevation (ft)} \\ \hline
\centering 1 \\ \centering(lowest) & \centering 50 (low) \newline 330 (high) & \centering 154 & 2,060 (low) \newline 7,356 (median) \newline 10,922 (high) \\ \hline
\centering  2 & \centering 331 (low) \newline500 (high) & \centering 153 & 3200 (low) \newline 8,123 (median) \newline 11,611 (high)  \\ \hline
\centering  3 & \centering  503 (low) \newline897 (high)  & \centering 153 & 3,060 (low) \newline 8,100 (median) \newline 11600 (high)  \\ \hline
\centering 4 \\ \centering(highest) & \centering 899 (low) \newline 2004 (high) & \centering 52  & 3,440 (low) \newline 6,205 (median) \newline 10,653 (high)  \\ \hline
\end{tabular}
\caption{Group-wise statistics based on Averaged Max SWE of locations across all training years. The groups are classified based on the 30th, 60th, and 90th percentile of max SWE values, and are agnostic to elevation. Groups 1 and 4 contain locations with the lowest and highest peak SWE values, respectively. }
\label{tab:locaton_groups}
\end{table*}
\begin{table*}[tphb]
\centering
\small
\setlength\tabcolsep{6pt}  % reduce column padding
\begin{tabular}{llccccc}
\toprule
\textbf{Metric} & \textbf{Model} & \textbf{2015} & \textbf{2016} & \textbf{2017} & \textbf{2018} & \textbf{2019} \\
\midrule
\multirow{2}{*}{NLL} & \fs & \textbf{8.33 $\pm$ 0.81} & \textbf{8.52 $\pm$ 0.82} & \textbf{8.46 $\pm$ 0.74} & \textbf{7.67 $\pm$ 0.62} & \textbf{7.15 $\pm$ 0.49} \\
                     & \rgp  & 15.04 $\pm$ 3.97 & 22.05 $\pm$ 6.70 & 29.01 $\pm$ 9.31 & 22.20 $\pm$ 6.86 & 23.93 $\pm$ 7.31 \\

\midrule
\multirow{2}{*}{ECE} & \fs & \textbf{0.16 $\pm$ 0.015} & \textbf{0.17 $\pm$ 0.017} & \textbf{0.21 $\pm$ 0.018} & \textbf{0.14 $\pm$ 0.017} & \textbf{0.17 $\pm$ 0.018} \\
                     & \rgp  & 0.41 $\pm$ 0.05 & 0.44 $\pm$ 0.04 & 0.47 $\pm$ 0.04 & 0.43 $\pm$ 0.05 & 0.46 $\pm$ 0.04 \\
\midrule
\multirow{2}{*}{Coverage} & \fs & \textbf{78.7 $\pm$ 1.56} & \textbf{77.1 $\pm$ 1.78} & \textbf{73.5 $\pm$ 1.87} & \textbf{80.1 $\pm$ 1.71} & \textbf{77.6 $\pm$ 1.86} \\
                          & \rgp  & 53.24 $\pm$ 5.2 & 50.82 $\pm$ 4.7 & 47.97 $\pm$ 4.6 & 51.59 $\pm$ 4.9 & 48.65 $\pm$ 4.9 \\
\bottomrule
\end{tabular}
\caption{Detailed qualities of uncertainty estimates produced by \fs{} and \rgp{} over the years. The mean and std of each metric is obtained from ten experiments. The reported metrics include NLL (Negative Log Likelihood), ECE (Expected Calibration Error), and Coverage.~Higher coverage and lower values for NLL and ECE indicate better uncertainty estimates.}
\label{tab:daily_gp_metric_std}
\end{table*}

% \begin{table*}[tb]
% \centering
% \small
% \setlength\tabcolsep{6pt}  % reduce column padding
% \begin{tabular}{llccccc}
% \toprule
% \textbf{Metric} & \textbf{Model} & \textbf{2015} & \textbf{2016} & \textbf{2017} & \textbf{2018} & \textbf{2019} \\
% \midrule
% \multirow{2}{*}{NLL} & \fs & \textbf{28.71} & \textbf{27.61} & \textbf{38.07} & \textbf{15.55} & \textbf{23.97} \\
%                      & \rgp  & 50.82 & 63.55 & 75.93 & 62.05 & 71.02 \\
% \midrule
% \multirow{2}{*}{ECE} & \fs & \textbf{0.67} & \textbf{0.60} & \textbf{0.64} & \textbf{0.53} & \textbf{0.59} \\
%                      & \rgp  & 0.68 & 0.70 & 0.70 & 0.70 & 0.72 \\
% \midrule
% \multirow{2}{*}{Coverage} & \fs & \textbf{27.82} & \textbf{34.80} & \textbf{30.30} & \textbf{41.61} & \textbf{35.68} \\
%                           & \rgp  & 26.76 & 24.08 & 24.30 & 24.35 & 22.62 \\
% \bottomrule
% \end{tabular}
% \caption{Weekly forecasting --- model performance over the years using NLL (Negative Log Likelihood), ECE (Expected Calibration Error), and Coverage metrics. These metrics show significant quality degradation from the daily forecasting and can be attributed to an inherent performance drop in a long forecasting horizon (4 weeks).}
% \label{tab:weekly_gp_metric}
% \end{table*}

\begin{table*}[thb]
\centering
\small
\setlength\tabcolsep{6pt}  % reduce column padding
\begin{tabular}{llccccc}
\toprule
\textbf{Metric} & \textbf{Model} & \textbf{2015} & \textbf{2016} & \textbf{2017} & \textbf{2018} & \textbf{2019} \\
\midrule
\multirow{2}{*}{NLL} & \fs & \textbf{25.49 $\pm$ 8.6} & \textbf{20.42 $\pm$ 6.1} & \textbf{35.07 $\pm$ 10.4} & \textbf{12.71 $\pm$ 3.3} & \textbf{20.10 $\pm$ 6.2} \\
                     & \rgp  & 26.05 $\pm$ 6.52 & 36.9 $\pm$ 8.87 & 46.92 $\pm$ 10.47 & 37.55 $\pm$ 8.75 & 41.07 $\pm$ 9.76 \\
\midrule
\multirow{2}{*}{ECE} & \fs & \textbf{0.61 $\pm$ 0.05} & \textbf{0.55 $\pm$ 0.05} & \textbf{0.62 $\pm$ 0.04} & \textbf{0.46 $\pm$ 0.06} & \textbf{0.53 $\pm$ 0.05} \\
                     & \rgp  & 0.62 $\pm$ 0.03 & 0.64 $\pm$ 0.028 & 0.66 $\pm$ 0.024 & 0.64 $\pm$ 0.025 & 0.67 $\pm$ 0.024 \\
\midrule
\multirow{2}{*}{Coverage} & \fs & \textbf{33.33 $\pm$ 5.2} & \textbf{39.88 $\pm$ 5.7} & \textbf{29.30 $\pm$ 4.9} & \textbf{48.38 $\pm$ 6.0} & \textbf{41.85 $\pm$ 5.5} \\
                          & \rgp  & 32.89 $\pm$ 3.0 & 30.20 $\pm$ 2.8 & 28.56 $\pm$ 2.3 & 30.08 $\pm$ 2.5 & 27.83 $\pm$ 2.4 \\
\bottomrule
\end{tabular}
\caption{Weekly forecasting --- model performance over the years using detailed NLL (Negative Log Likelihood), ECE (Expected Calibration Error), and Coverage metrics. The mean and std of
each metric is obtained from ten experiments.These metrics show significant quality degradation from the daily forecasting and can be attributed to an inherent performance drop in a long forecasting horizon (4 weeks).}
\label{tab:weekly_gp_metric}
\end{table*}
% \begin{table}[thb]
% \centering
% \colorbox{darkgrn!10}{
% \begin{tabular}{|l|c|}
% \hline
% \textbf{Model} & \textbf{Parameter Count (in millions)} \\
% \hline
% ForeSWE & 426 \\
% \hline
% NVA-Base & 400 \\
% \hline
% Tp-Att & 200 \\
% \hline
% Sp-Att & 155 \\
% \hline
% Transformer & 12 \\
% \hline
% LSTM & 4 \\
% \hline
% TFT & 30 \\
% \hline
% \end{tabular}
% }
% \caption{Parameter count of models compared in this work.}
% \label{tab:param_counts}
% \end{table}

\begin{table}[htbp]
\centering
\begin{tabular}{lcccc}
\hline
\textbf{Model} & \textbf{Spatial} & \textbf{Temporal} & \textbf{Attribute Interaction} & \textbf{Params (M)} \\
\hline
Sp-Att      & X &   &   & 155 \\
% \hline
Tp-Att      &   & X &   & 200 \\
% \hline
Transformer &   & X &   & 12 \\
% \hline
LSTM        &   & X &   & 4 \\
% \hline
TFT         &   & X &   & 30 \\
% \hline
NVA-Base    & X & X &   & 400 \\
% \hline
Raw-GP      & X & X &   & 220 \\ 
% \hline
\textbf{ForeSWE} & X & X & X & \textbf{426} \\
\hline
\end{tabular}
\caption{Model characteristics: spatial, temporal, and temporal-attribute interaction, with their respective parameters count.}
\label{tab:model_param}
\end{table}

\begin{table}[h!]
\centering
\begin{tabular}{lccc}
\hline
\textbf{Component} & \textbf{Learning Rate} & \textbf{Epochs} & \textbf{Dimension} \\
\hline
\fs{} (Attention Part) & 1e-2,\ 5e-4,\ 5e-5 & 5,\ 8,\ 20 & 512,\ 1024 \\
\fs{} (GP Part)       & 1e-3, \ 1e-2, \ 1.2e-1      & 2,\ 4,\ 10  & 8,\ 16 \\
\hline
\end{tabular}
\caption{Hyperparameter search space for \fs{} attention and GP components. The final hyperparameter set was chosen based on NSE value in both daily and weekly forecasting.}
\label{tab:foreswe_hyperparams}
\end{table}

\begin{table}[thb]
\centering
\begin{tabular}{lccc}
\hline
\textbf{Model} & \textbf{Learning Rate} & \textbf{Epochs} & \textbf{Dimension} \\
\hline
\spatt      & 5e-4   & 10 & 1024 \\
\tpatt       & 5e-4   & 10 & 1024 \\
\tar   & 1e-3    & 8  & 1024 \\
\lstm          & 5e-4   & 10 & 1024 \\
\tft      & 5e-4   & 1 & 1024 \\
\nva      & 5e-5     & 10 & 1024 \\
\rgp       & 0.12     & 4  & 21 \\
\fs       & 5e-4 (attn) \& 0.12(GP)   & 8  & 1024 (attn) \& 8 (GP) \\
\hline
\end{tabular}
\caption{Model final hyperparameters and dimensionality. These hyperparameters are chosen based on the NSE of
prediction in both daily and weekly forecasting.}
\label{tab:model_hyperparams}
\end{table}
\captionsetup{font={normalsize}}
\begin{figure*}[tbh]
\includegraphics[scale=0.55]{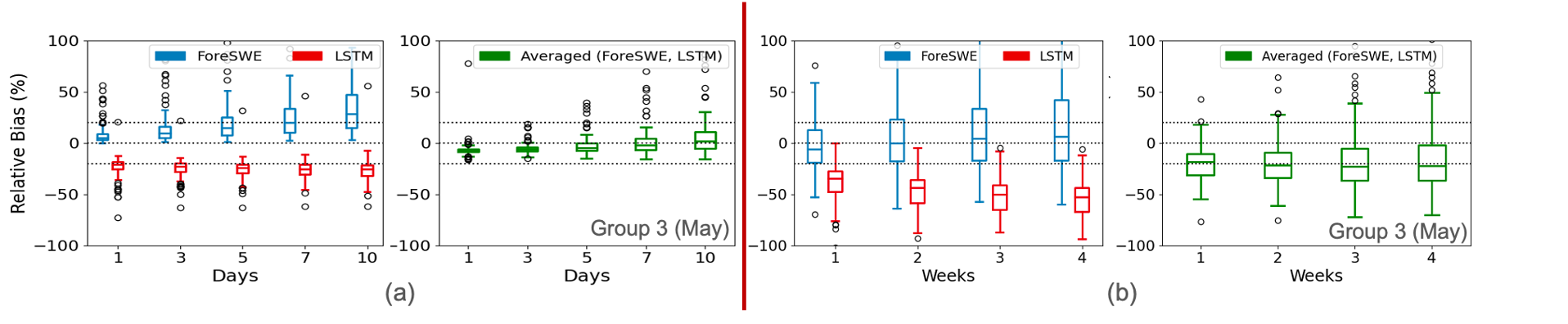}
\captionof{figure}{\normalsize (a) Group 3 daily relative bias with \fs~ and \lstm~ model for May. (b) Group 3 weekly relative bias with \fs~ and \lstm~ model for May. The dotted lines show the reference of relative bias between $\pm$ 20\%. \textbf{Findings} --- \lstm~ model has consistent underprediction in its daily and weekly forecasts. Therefore, it is better to combine \fs~ with \lstm~ only when it has significant overprediction or else can lead to reduced performance. For example, in (b) the combination pushed the median relative bias to -\%20, even though it reduced the spread of the distribution.
}
\label{fig:group_3_may}
\end{figure*}
\captionsetup{font={normalsize}}
\begin{figure*}[tbh]
    \centering
    \includegraphics[scale=0.70]{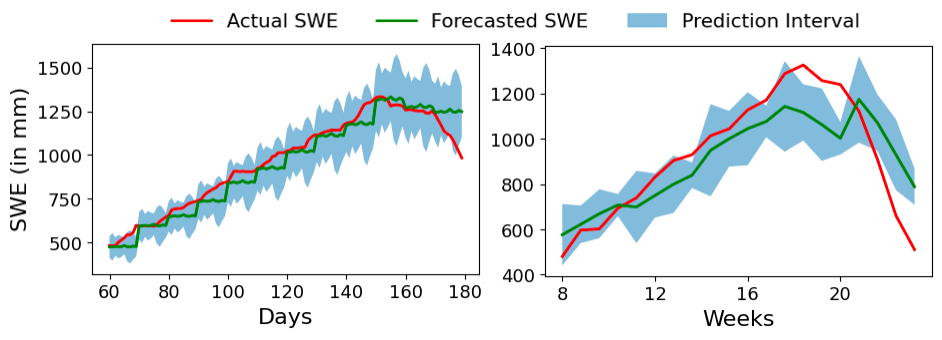}\vspace{-2mm}
    \caption{\normalsize \fs{} --- Prediction intervals for daily and weekly SWE forecasting starting Feb 1 (60 days or 8 weeks from Dec 1).}
    \label{fig:gpimage}
\end{figure*}
\captionsetup{font={normalsize}}
\begin{figure*}[tbh]
    \centering
    \includegraphics[scale=0.85]{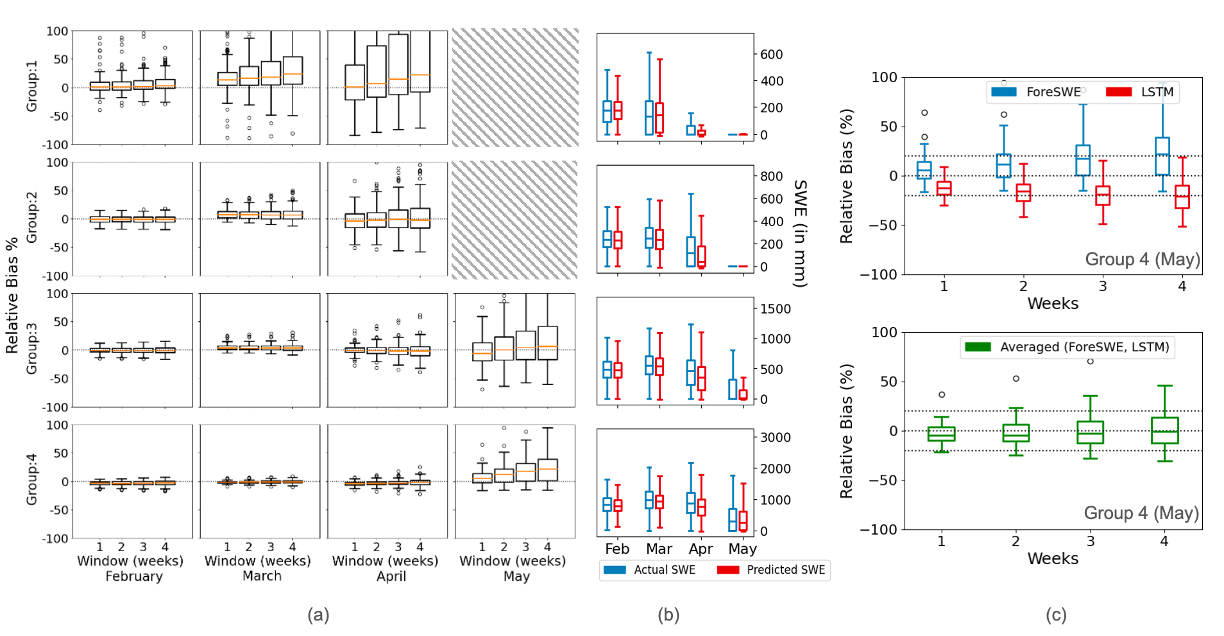}
    
    \caption{\normalsize \textbf{Weekly forecasting} --- (a) \fs~ model's  relative bias, by the different months and location groups (with forecasting horizons ranging from 1 to 4 weekly points). Groups 1 and 2 are blocked for May as the snow has melted completely at these locations. (b) Actual and \fs{} Predicted SWE availability in different groups across the active SWE months. (c) (upper) Relative bias of \fs~ model against a temporal model (\lstm) for Group 4 in May over different forecasting windows. (lower) Relative bias with the ensembling of \fs~ and \lstm~ for Group 4 in May. The dotted lines in the plot mark $\pm$ 20\%. \textbf{Findings} --- The relative bias of ~\fs~ is consistently better in accumulation months for all location groups. However, during the period of rapid melt, which varies across location groups, the short-term forecast (1-2 weeks) has good performance and increases overestimation with increasing horizon. In May, the overestimation in \fs~ can be mitigated when combined with \lstm~ model, which has consistent underprediction.} 
    \label{fig:weekly_fgp_months}
\end{figure*}
\captionsetup{font={normalsize}}
\begin{figure*}[tbh]
\centering
\includegraphics[scale=0.55]{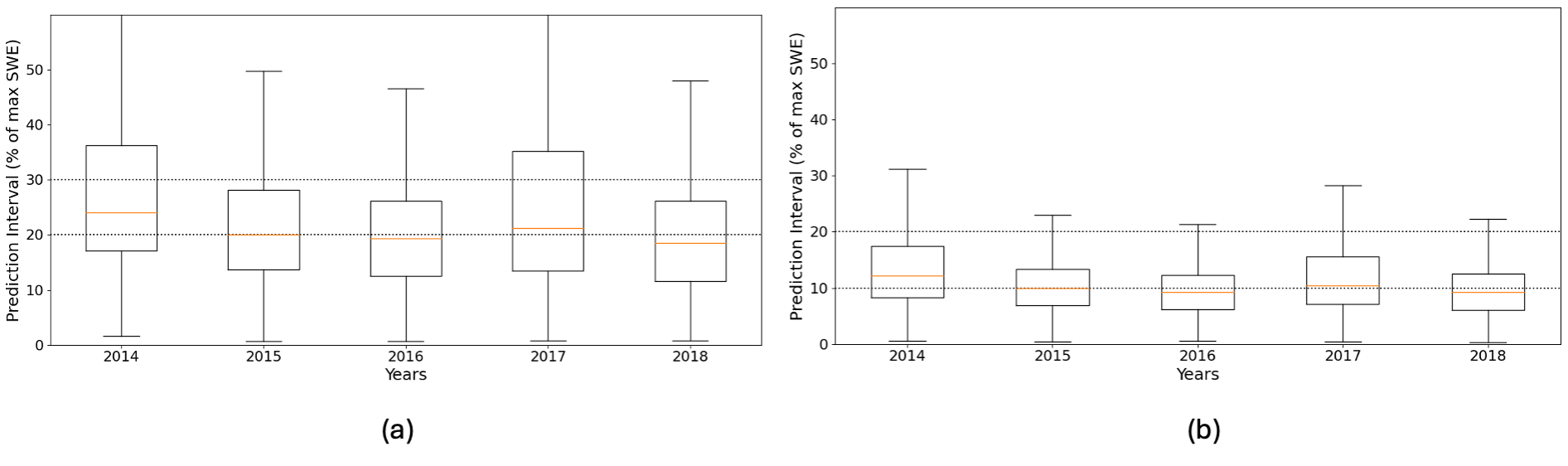}
\captionof{figure}{\fs{} \normalsize (a) Daily forecasts. (b) Weekly forecasts. (Prediction interval (\%) in terms of max SWE for all locations across all test years. \textbf{Findings} --- \fs's confidence estimates lie between 10-30\% of the max SWE value for each location, consistently across all test years. Additionally, the prediction interval of weekly forecasting is narrower than daily forecasting, leading to high confidence in its prediction and reduced confidence interval calibration.
}
\label{fig:pred_interval_info}
\end{figure*}

\end{document}